\documentclass[letterpaper]{article} 
\usepackage{aaai23}  
\usepackage{times}  
\usepackage{helvet}  
\usepackage{courier}  
\usepackage[hyphens]{url}  
\usepackage{graphicx} 
\usepackage{natbib}  
\usepackage{caption} 
\usepackage{algorithm}
\usepackage{algorithmic}

\usepackage{newfloat}
\usepackage{listings}
\DeclareCaptionStyle{ruled}{labelfont=normalfont,labelsep=colon,strut=off} 
\floatstyle{ruled}
\newfloat{listing}{tb}{lst}{}
\floatname{listing}{Listing}
\usepackage{amsmath,amssymb} 
\usepackage{gensymb}
\usepackage[symbol]{footmisc}

\usepackage[capitalize]{cleveref}
\crefname{section}{Sec.}{Secs.}
\Crefname{section}{Section}{Sections}
\Crefname{table}{Table}{Tables}
\crefname{table}{Tab.}{Tabs.}
\usepackage{multirow}
\usepackage{booktabs}
\usepackage{tabularx}
\usepackage{makecell}
\usepackage{subcaption}
\usepackage{nicematrix}
\usepackage{xcolor,colortbl}

\newcommand{\norm}[1]{\left\lVert#1\right\rVert} 
\newcommand{\valseen}{Val-Seen{ }}  
\newcommand{\valunseen}{Val-Unseen{ }}
\newcommand{\vlnbert}{VLN$\protect\circlearrowright$BERT{}}

\newcommand{\videourl}{\url{https://youtu.be/4fVG0vg7yXI}}

\definecolor{Gray}{gray}{0.95}
\newcolumntype{a}{>{\columncolor{Gray}}c}
\newcolumntype{b}{>{\columncolor{white}}c}

\title{Simple and Effective Synthesis of Indoor 3D Scenes}
\author{
    Jing Yu Koh,\textsuperscript{\equalcontrib,\rm 1, \footnote{Now at Carnegie Mellon University.}}
    Harsh Agrawal,\textsuperscript{\equalcontrib,\rm 2, \footnote{Work done while at Google.}} \\
    Dhruv Batra,\textsuperscript{\rm 2} Richard Tucker,\textsuperscript{\rm 1} Austin Waters,\textsuperscript{\rm 1} Honglak Lee,\textsuperscript{\rm 3}  Yinfei Yang,\textsuperscript{\rm 4, $^\ddagger$} \\
    Jason Baldridge,\textsuperscript{\rm 1} Peter Anderson\textsuperscript{\rm 1}
}
\affiliations{
    \textsuperscript{\rm 1} Google Research 
    \textsuperscript{\rm 2} Georgia Institute of Technology
    \textsuperscript{\rm 3} University of Michigan
    \textsuperscript{\rm 4} Apple
}

\usepackage{bibentry}

\begin{document}

\maketitle

\begin{abstract}
We study the problem of synthesizing immersive 3D indoor scenes from one or a few images. 
Our aim is to generate high-resolution images and videos from novel viewpoints, including viewpoints that extrapolate far beyond the input images while maintaining 3D consistency. 
Existing approaches are highly complex, with many separately trained stages and components. 
We propose a simple alternative: an image-to-image GAN that maps directly from reprojections of incomplete point clouds to full high-resolution RGB-D images.
On the Matterport3D and RealEstate10K datasets, our approach significantly outperforms prior work when evaluated by humans, as well as on FID scores. Further, we show that
our model is useful for generative data augmentation. A vision-and-language navigation (VLN) agent trained with trajectories spatially-perturbed by our model 
improves success rate by up to 1.5\% over a state of the art baseline on the mature R2R benchmark. Our code is publicly released to facilitate generative data augmentation and applications to downstream robotics and embodied AI tasks.
\end{abstract}

\section{Introduction} \label{sec:intro}

We synthesize immersive 3D indoor scenes from one or more context images captured along a trajectory. Our aim is to generate high-resolution images and videos 
from novel viewpoints, including viewpoints that extrapolate far beyond the context image(s).

Solving this problem would make photos 
and videos 
interactive and immersive, with applications not only to content creation but also robotics and embodied AI. For example, models that can predict around corners could be used by navigation agents as world models~\cite{ha2018world} for model-based planning in novel environments~\cite{koh2021pathdreamer,finn2017deep}. Such models could also be used to train agents in interactive environments synthesized from static images or video.

\begin{figure*}[t]
    \centering
    \includegraphics[width=1.0\linewidth]{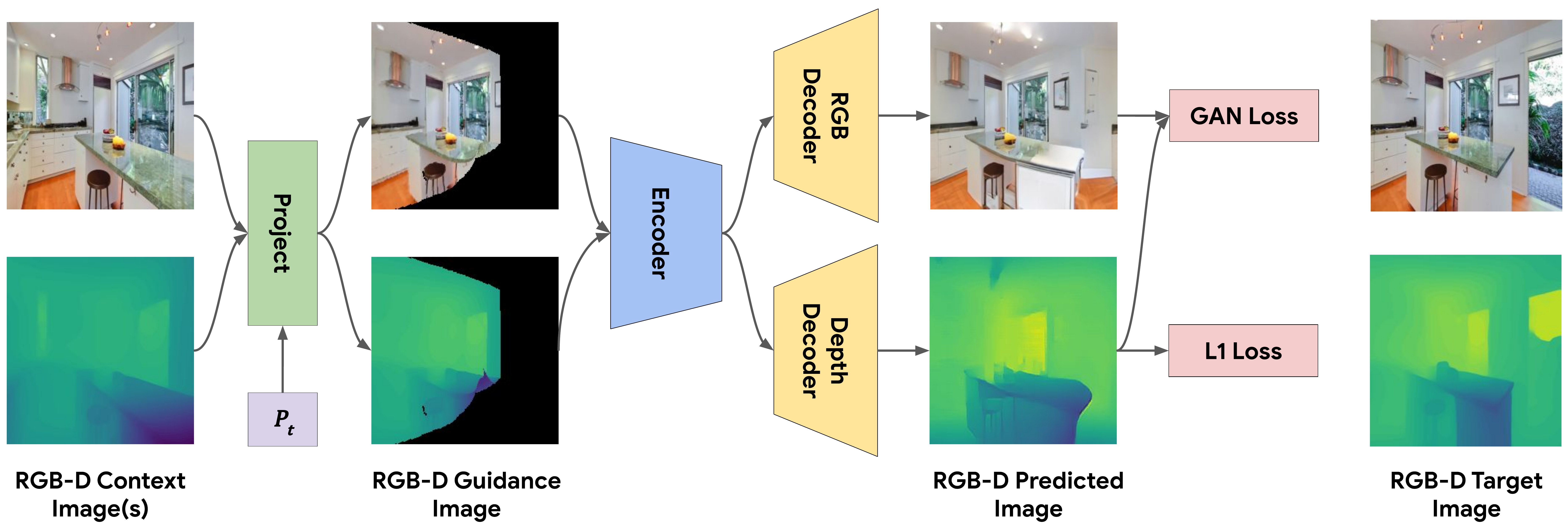}
    \caption{Our lightweight approach to 3D scene synthesis accumulates context images in an RGB point cloud. To generate a new viewpoint, we simply apply an image-to-image GAN to the \textit{guidance image} of the reprojected point cloud. We achieve surprisingly strong results with this approach, significantly outperforming more complicated models.}
    \label{fig:model}
\end{figure*}

Previous approaches attempting this under large viewpoint changes~\cite{wiles2020synsin,koh2021pathdreamer,rockwell2021pixelsynth} typically operate on point clouds, which are accumulated from the available context images. The use of point clouds naturally incorporates camera projective geometry into the model and helps maintain the 3D consistency of the scene~\cite{mallya2020world}.
To generate novel views, these approaches reproject the point-cloud relative to the target camera pose into an RGB-D \textit{guidance image} (see \cref{fig:model}). These guidance images are sparse because of missing context and completing them requires extensive inpainting and outpainting. In prior work, Pathdreamer~\cite{koh2021pathdreamer} achieves this by assuming the availability of semantic segmentations, and combining a stochastic depth and semantic segmentation (structure) generator with an RGB image generator infused with semantic, depth and guidance information via Multi-SPADE spatially adaptive normalization layers~\cite{park2019semantic,mallya2020world}. On the other hand, PixelSynth~\cite{rockwell2021pixelsynth} creates guidance images using a differentiable point cloud renderer, generates a support set of additional views using PixelCNN++~\cite{salimans2017pixelcnn++} operating on the latent space of a VQ-VAE~\cite{vqvae2}, combines and refines these images using a GAN~\cite{goodfellow2014generative} similar to SynSin~\cite{wiles2020synsin}, and then repeats this process over many samples using a combination of discriminator loss and the entropy of a scene classifier to select the best output.

We propose a simple alternative: an image-to-image GAN that maps directly from guidance images to high-resolution photorealistic RGB-D (see \cref{fig:model}).
Compared to Pathdreamer, our simple model forgoes the stochastic structure generator, spatially adaptive normalization layers, dependence on semantic segmentations, and multi-step training. 
By dropping the dependence of semantic segmentation inputs, we unlock training on a much broader range of data, such as more commonly available RGB-D datasets~\cite{xia2018gibson,li2018megadepth,silbermanNYUDepth}, and video data such as the RealEstate10K dataset~\cite{zhou2018stereo} from YouTube.
We eschew many components of PixelSynth: differentiable rendering, support set generation using PixelCNN++and a VQ-VAE, and the multiple sampling and re-ranking procedure.

Perhaps surprisingly, with random masking of the guidance images during training, plus other architectural changes supported by thorough ablation studies, our lightweight approach outperforms prior work. In human evaluations of image quality, our model is preferred to Pathdreamer in 60\% of comparisons and preferred to PixelSynth in 77\%. Our FID scores on 360\degree{} panoramic images from Matterport3D~\cite{Matterport3D} improve over Pathdreamer's by 27.9\% relatively (from 27.2 to 19.6) when predicting single step viewpoint changes (an average of 2.2m), and from 65.8 to 58.0 when predicting over longer trajectories containing a sequence of 6 novel viewpoints. On RealEstate10K~\cite{zhou2018stereo} -- a collection of real estate video walkthroughs from YouTube -- our FID scores outperform PixelSynth, improving from 25.5 to 23.5 (and from 23.6 to 21.5 for indoor images).  Our model is capable of composing compelling 3D synthetic environments from a single image, which can be used to produce realistic video renderings.\footnote{Video results: \videourl.}

Motivated by the strong results on image generation, we also show the usefulness of our model for data augmentation in embodied AI. We focus on the vision-and-language navigation (VLN) task, which requires an agent to follow natural language navigation instructions in previously unseen photorealistic environments. Training data for the task consists of instruction-trajectory demonstrations, where each trajectory is defined by a sequence of high-resolution 360\degree{} panoramas. Using our model, we spatially perturb the location of the training panoramas by synthesizing views up to 1.5m away and augment the training dataset. This reduces overfitting to the incidental details of these trajectories, and improves the success in unseen environments by an additional 1\% on its own, or 1.5\% when combined with renders of spatially-perturbed images from the Habitat~\cite{habitat19iccv} simulator -- achieving state-of-the-art performance on the R2R test set (\cref{table:vln_aug_results}). Our code is publicly released\footnote{\url{https://github.com/google-research/se3ds}} to facilitate generative trajectory augmentation and applications to downstream tasks.

\section{Related Work}

\paragraph{Novel View Synthesis.}
Synthesizing novel views from sets of 2D images has been studied extensively from the lens of multi-view geometry~\cite{debevec1996modeling,avidan1997novel,zitnick2004high}, and more recently with deep learning based approaches~\cite{flynn2016deep,kar2017learning,henzler2018single,flynn2019deepview,srinivasan2019pushing,zhou2018stereo,mildenhall2019local}. Various methods using explicit 3D scene representations have been proposed, including point cloud representations~\cite{wiles2020synsin}, layered depth images~\cite{dhamo2019peeking}, and mesh representations~\cite{shih2020photography}. More recently, Neural Radiance Fields (NeRF)~\cite{mildenhall2020nerf} models have achieved impressive results on novel view synthesis. NeRFs optimize an underlying continuous volumetric scene function, learning an implicit 3D scene representation from multiple images. Follow up papers have extended the NeRF formulation to large-scale environments~\cite{tancik2022block}, unconstrained photo collections~\cite{martinbrualla2020nerfw}, learning scene priors~\cite{yu2020pixelnerf}, using multi-scale representations~\cite{barron2021mip}, and reducing the amount of input images required~\cite{jain2021putting}. However, at present NeRF models primarily focus on scene representation rather than generalization to unseen environments, and are unable to perform realistic synthesis of previously unseen 3D scenes at high resolution (which is our focus). 

While early work in novel view synthesis focused on smaller view changes and settings with multiple input images, recent methods tackle single-image novel view synthesis \cite{tucker2020single,hu2021worldsheet,shih2020photography} and large viewpoint changes \cite{wiles2020synsin,koh2021pathdreamer,rockwell2021pixelsynth,rombach2021geometry} and long-term future prediction for indoor scenes \cite{ren2022look}. The primary challenge of this task is being able to handle both inpainting of missing pixels, as well as outpainting of large regions of the image from limited context, while maintaining consistency with the existing scene. The closest works to ours are PixelSynth~\cite{rockwell2021pixelsynth} and Pathdreamer~\cite{koh2021pathdreamer} (see introduction).

\paragraph{Point Cloud Rendering.}
A number of previous works explore rendering novel views from point clouds. 
The Gibson simulator~\cite{xia2018gibson} combined point cloud rendering with a neural net `filler' to fix artifacts and produce more realistic images, although the neural net was trained with perceptual loss~\cite{johnson2016perceptual} rather than GAN based losses. Several papers~\cite{yu2020point,fu20203d} propose methods to inpaint object point clouds with new points in order to fill holes, which would allow more realistic images can be synthesized. \citet{song2020deep} proposes a model which takes a colored 3D point cloud of a scene as input, and synthesizes a photo-realistic image from a novel viewpoint. 
Whereas,
\citet{cortinhal2021semantics} relies only on the semantics of a scene to synthesize a panoramic color image from a given full 3D LiDAR point cloud. 
These methods often consider smaller point clouds, which limits their efficiency and viability for high resolution imagery. For example, \cite{song2020deep} trains on point clouds of size $4096\times6$, while the accumulated point cloud for our method uses RGB-D images of size $1024\times512$ (at least $64\times$ larger).

\paragraph{Generative Data Augmentation.}
Generative models are good alternatives to standard data augmentation approaches. They can synthesize artificial samples that match the distribution and characteristics of an underlying dataset, augmenting the training set with additional novel examples~\cite{antoniou2017data,sandfort2019data}. We examine whether novel view synthesis can be used to create new training trajectories for a navigation agent, by spatially-perturbing camera viewpoints \textit{post hoc}. 

In imitation learning, expert demonstrations must be augmented to minimize the differences between the state distribution seen in training, and those induced by the agent during inference, which otherwise causes compounding errors~\cite{ross2010efficient}. In practice this usually means augmenting the dataset with examples of recoveries from error, which may be rare in expert demonstrations. For example, robots and self-driving vehicles~\cite{codevilla2018end,bojarski2016end} can be instrumented to record from three cameras simultaneously (one facing forward and two shifted to the left and right). This allows recordings from the shifted cameras, as well as intermediate synthetically reprojected views, to be added to the training set with adjusted control signals to simulate recovery from drift~\cite{codevilla2018end,bojarski2016end}. In effect, we propose a flexible, hardware-free alternative to the multi-camera setup. We investigate this in the context of indoor vision-and-language navigation (VLN) on the R2R dataset~\cite{anderson2018vision}. To the best of our knowledge, we are the first to show benefits from data augmentation using novel view synthesis models in a photorealistic setting.

\section{Approach} \label{sec:approach}

We aim to synthesize high-resolution images and videos from novel viewpoints in buildings, conditioning on one or more RGB and depth (RGB-D) observations as context. Specifically, given context consisting of a sequence of RGB-D image observations and their associated camera poses $(I_{1:t-1}, P_{1:t-1})$, our goal is to generate realistic RGB-D images for one or more target camera poses $[P_t, P_{t+1}, \cdots, P_T]$. Target poses may require extrapolating far beyond the context images (e.g., predicting around corners), requiring the model to generate and in-fill potentially large regions of missing information -- even entire rooms.

\paragraph{Depth Estimation and Guidance Images.}
Our model requires depth values and camera poses for the context images $(I_{1:t-1}, P_{1:t-1})$ to create an accumulated point-cloud. Given a new pose $P_t$, we reproject our accumulated point cloud, similar to \cite{mallya2020world,liu2021infinite,koh2021pathdreamer} into an image from that viewpoint. We call that image a \emph{guidance image} because it will be used to guide the inpainting and outpainting process later. 
In our experiments, we report results using both ground-truth and estimated depth values and camera poses, depending on the dataset. 
When predicting over multiple steps $[P_t, P_{t+1}, \cdots, P_T]$, \emph{predictions} are accumulated in the point cloud to maintain 3D consistency. We do not make assumptions about the input image format, which enables our model to run on equirectangular panoramas from Matterport3D~\cite{Matterport3D} as well as perspective images from RealEstate10K~\cite{zhou2018stereo}. To accommodate both formats, we simply use the appropriate camera models (while keeping model architecture the same).

\paragraph{Model Architecture.} \label{sec:model_architecture}
We propose a simple, single-stage, end-to-end trainable model to convert a \textit{guidance image} directly into a high-resolution photorealistic RGB-D output. \cref{fig:model} provides an overview of this process. Our model uses an encoder-decoder CNN architecture inspired by RedNet~\cite{jiang2018rednet}. We use ResNet-101~\cite{he2016deep} as the encoder, and a `mirror image' of the ResNet-101 as the decoder, replacing convolutions with transposed convolutions for upsampling. A single encoder is used for both the RGB and depth inputs in the guidance, but separate decoders are used for predicting RGB and depth outputs. Following RedNet, skip connections are introduced between the encoder and decoders to preserve spatial information. 

Inpainting or outpainting large image regions that lack input guidance is a major challenge, e.g. predicting around corners or under large viewpoint changes. To overcome this, we replace all convolutions in the encoder with partial convolutions~\cite{liu2018image}. Partial convolutions only convolve
valid regions in the input with the convolution kernels, so the output features are minimally affected by missing pixels in the guidance images. Further, we increase the effective receptive field size by adding four additional convolutional layers to the output of the encoder helping propagate non-local information. We found this to be essential, without which the model does not learn anything meaningful.

We also experimented with the SVG~\cite{denton2018stochastic} structure generator proposed in Pathdreamer~\cite{koh2021pathdreamer}. This module learns a prior distribution to model stochasticity in the predictions. During training, the prior distribution is learnt by minimizing KL-divergence from the posterior distribution derived from ground truth data. At inference time, different outputs are generated by sampling from the prior distribution. We found that generation quality was worse when this module was included (as measured by FID and visual inspection), despite improvement in image diversity (more details in the \textit{Ablation Studies} section). Hence, we omit the SVG module in our final model for the sake of simplicity and performance. However, we stress that this module can be easily incorporated to enable a trade-off between generation quality and diversity.

\paragraph{Loss Functions.}
The encoder-decoder model $G$ is optimized to minimize a joint loss consisting of an $L_1$ loss between the predicted depth $\hat{d}_t$ and the ground-truth depth $d_t$, and an adversarial loss on depth and RGB images. The discriminator $D$ used for adversarial training is based on PatchGAN~\cite{pix2pix2017}, and applied to the 4-channel RGB-D image (either generated $[\hat{I}_t, \hat{d}_t]$ or ground-truth $[I_t, d_t]$). Note that while the use of separate decoders could allow for greater divergence between the RGB and depth predictions, this is mitigated by the discriminator which helps to enforce consistency between RGB and depth, which is essential for multi-step predictions.
The training losses for the generator and discriminator are:
\begin{align} \label{eq:training_loss}
\mathcal{L}_{G} =& -\lambda_{\text{GAN}}~ \mathbb{E}_{x_t}\Big [D\big(G(I_{1:t-1}, T_{1:t})\big)\Big ]  + \lambda_{\text{D}} \norm{\hat{d}_t - d_t}_1  \nonumber\\ 
\mathcal{L}_D=&-\mathbb{E}_{x_t}\Big[\min\big(0, -1 + D(I_t)\big)\Big] \nonumber\\
&- \mathbb{E}_{x_t}\Big[\min\big(0, -1- D(G(I_{1:t-1}, T_{1:t}))\big)\Big]
\end{align}

Notably, we avoid the VGG perceptual loss commonly used in prior work on conditional image synthesis~\cite{park2019semantic,koh2021pathdreamer,mallya2020world} as we find that it is unnecessary for strong performance. We justify this through careful ablation experiments.

In all experiments we train the model end-to-end from scratch, and we randomly mask up to 75\% of the input guidance image for data augmentation. Masking has been shown to be an effective method of pre-training visual representations~\cite{he2021masked,dosovitskiy2020image}. We similarly find that this masking strategy significantly improves the generation quality of our model in unseen environments.

\begin{figure*}[t]
    \centering
    \includegraphics[width=1.0\linewidth]{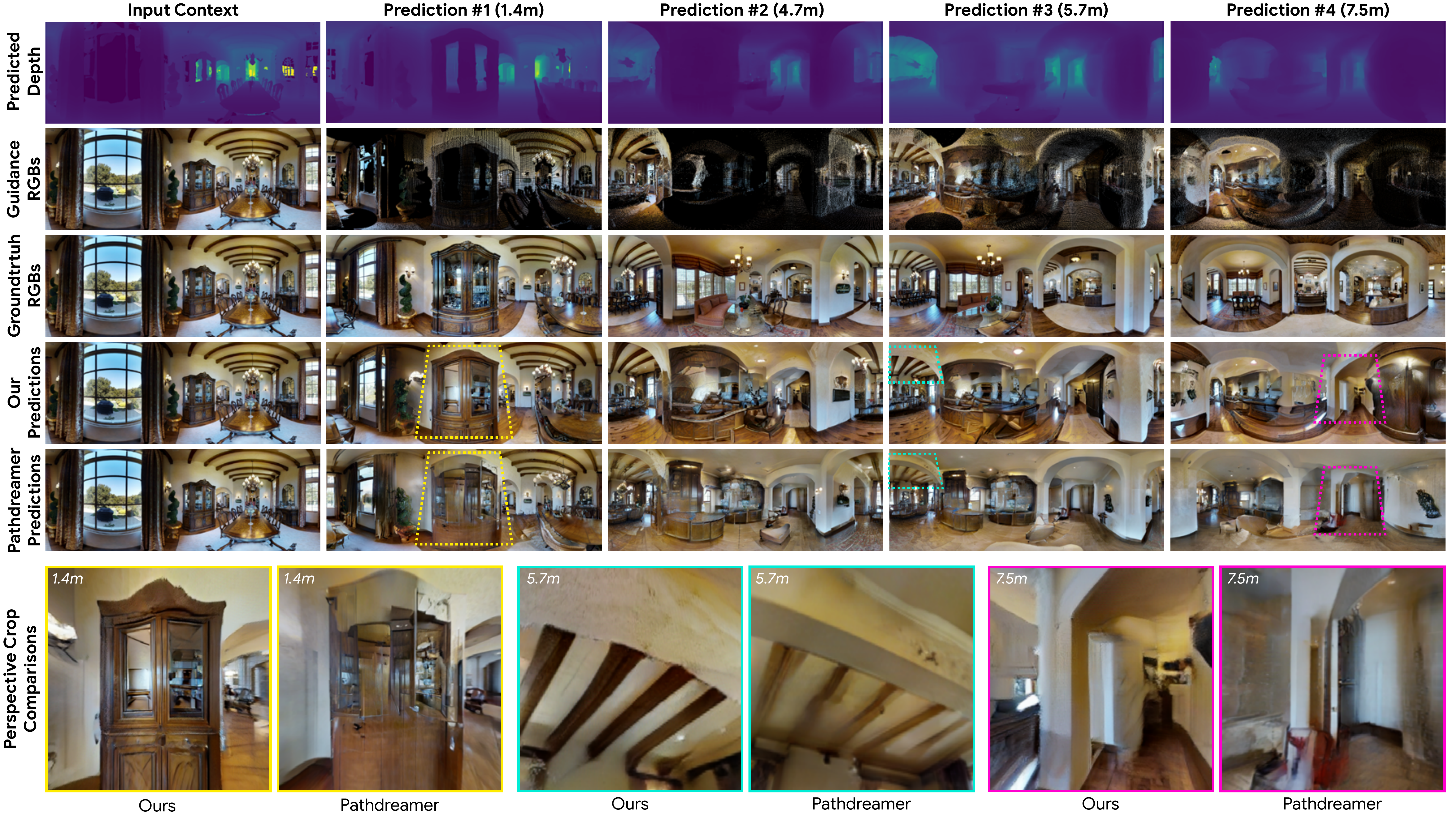}
    \caption{Qualitative example of a prediction sequence in an unseen Matterport3D building. One RGB-D pano is provided as context, and a sequence of panos at novel viewpoints (up to 7.5m away) are generated. Our predictions are clearer than Pathdreamer~\cite{koh2021pathdreamer}
    -- in last row, our model produces more realistic images of the wooden cabinet at 1.4m, the ceiling rafters at 5.7m, and the passageway 7.5m away.}
    \label{fig:qualitative_example}
\end{figure*}

\section{View Synthesis Experiments}

We conduct experiments on two datasets of diverse indoor environments: Matterport3D~\cite{Matterport3D}, which contains 3D meshes of 90 buildings reconstructed from 11K high-resolution RGB-D panoramas (panos), and RealEstate10K~\cite{zhou2018stereo}, a collection of up to 10,000 YouTube video walkthroughs of real estate properties. Few prior works attempt view synthesis from a single image under large viewpoint changes. We compare to PixelSynth~\cite{rockwell2021pixelsynth}, which builds on and outperforms SynSin~\cite{wiles2020synsin}, and Pathdreamer~\cite{koh2021pathdreamer}. We report automated Fréchet Inception Distance (FID)~\cite{heusel2017gans} scores (lower is better) and pairwise human evaluations of image quality. 

\subsection{Matterport3D Experiments} \label{sec:matterport_results}

\begin{table}[t]
\begin{center}
\setlength{\tabcolsep}{3pt}
\scriptsize
\resizebox{1.0\linewidth}{!}{%
\begin{tabular}{lccccccccaa}
\noalign{\smallskip}
 & & \multicolumn{3}{c}{\textbf{Inputs}} & & \multicolumn{2}{c}{\textbf{\valseen}} & & \multicolumn{2}{a}{\textbf{\valunseen}} \\
\cmidrule{3-5} \cmidrule{7-8} \cmidrule{10-11}
\textbf{Model} & \textbf{Context} & Seg & RGB & Depth & & 1 Step & 1--6 Steps & & 1 Step & 1--6 Steps \\ \midrule
Pathdreamer & 1 & \checkmark & \checkmark & \checkmark & & 26.2 & 41.7 & & 34.8 & 70.4 \\
Pathdreamer$^{\dagger}$ & 1 & \checkmark & \checkmark & \checkmark & & 20.4 & \textbf{36.0} & & 27.2 & 65.8 \\
Ours & 1 & - & \checkmark & \checkmark & & \textbf{19.4} & 55.4 & & \textbf{19.6} & \textbf{58.0} \\
\midrule
Pathdreamer & 2 & \checkmark & \checkmark & \checkmark & & 25.8	& 38.4 &  & 38.2 &	61.0 \\
Pathdreamer$^{\dagger}$ & 2 & \checkmark & \checkmark & \checkmark & & 19.8 & \textbf{32.6} & & 31.4 & 55.8 \\
Ours & 2 & - & \checkmark & \checkmark & & \textbf{19.3} & 45.3 & & \textbf{22.4} & \textbf{49.5} \\
\midrule
Pathdreamer & 3 & \checkmark & \checkmark & \checkmark & & 25.6 & 36.7 & & 38.5 & 52.9 \\
Pathdreamer$^{\dagger}$ & 3 & \checkmark & \checkmark & \checkmark & & 19.6 & \textbf{30.2} & & 32.0 & 47.1 \\
Ours & 3 & - & \checkmark & \checkmark & & \textbf{19.0} & 39.0 & & \textbf{22.3} & \textbf{39.9} \\
\midrule
\end{tabular}
}
\caption{FID scores ($\downarrow$) for generated RGB sequences on R2R paths in Matterport3D, using 1--3 context images. On Val-unseen, our model outperforms Pathdreamer 
in all settings. $^{\dagger}$ indicates restated results excluding blurred regions (directly comparable to ours).}
\label{table:mp3d_results}
\end{center}
\end{table}

Both PixelSynth and Pathdreamer report results on Matterport3D. However, as the two papers are concurrent work, the evaluation procedures are not comparable. PixelSynth is trained and evaluated on $256\times256$ perspective renders from the reconstructed 3D meshes. These renders are not photorealistic and often contain large regions of missing/black pixels due to poor reconstruction. Pathdreamer is trained and evaluated on the high-res $1024\times512$ 360\degree{} panoramic images, and evaluated over multiple steps of prediction. Therefore, on the Matterport3D dataset we follow the more challenging Pathdreamer evaluation procedure. 

\paragraph{Training and Evaluation.} Following Pathdreamer, we train our model using $1024 \times 512$ equirectangular RGB-D images and ground-truth pose information. Unlike Pathdreamer, our model does not require ground-truth semantic segmentations as input, and we train for only single-step prediction, i.e., predicting the pano at an adjacent viewpoint using one context pano as input. Evaluations are based on 

\valseen and \valunseen splits of the Room-to-Room (R2R) dataset~\cite{anderson2018vision}, which are comprised of sequences of adjacent panoramas ($\sim$2.2m apart). Given the first RGB-D pano in the path and its $(x, y, z)$ pose as context, the model must generate panos for the remainder of the path given only their poses (up to a maximum of 6 steps). 

\paragraph{FID Scores.} In \cref{table:mp3d_results} we report FID scores for the generated RGB images over 1--6 prediction steps (representing trajectory rollouts of 2--13m), using 1--3 panos as context. We evaluate in two settings -- novel viewpoints in environments seen during training (Val-Seen) and previously unseen (Val-Unseen) environments. Since the top and bottom 12.5\% of each Matterport3D pano is blurred, we crop these areas before calculating FID and re-state the previously reported results from Pathdreamer for fair comparison. 
On Val-Unseen, based on FID score, our model outperforms Pathdreamer in all settings. When using a single context image for 1-step prediction,  we improve FID scores from 27.6 to 19.6. Similarly, when using 3 panos as context, FID scores improve from 32.0 to 22.3. We notice similar gains for multi-step (1-6 steps) prediction. On Val-seen, we improve FID scores slightly over Pathdreamer but perform worse on multi-step prediction in Val-Seen environments.
We attribute this to Pathdreamer's recurrent training regime for their Structure Generator, under which the model is trained over multiple prediction steps while accumulating its own outputs as additional context. In initial experiments we found that this improved results in the training environments, but did not generalize to unseen environments, perhaps because the accumulated context from the model predictions doesn't reconcile with the target image used in the $L_1$ depth reconstruction loss. In unseen environments (which is our focus), our model performs better on FID score at every step. As shown in \cref{fig:qualitative_example}, despite using a far simpler model than Pathdreamer that does not require ground-truth semantic segmentation inputs, our model produces a clearer image of the wooden cabinet at 1.4m, the wooden ceiling rafters at 5.7m, and the passageway at 7.5m away.

\paragraph{Human evaluations.} We perform human evaluations of 1,000 image pairs from our model and Pathdreamer. Each pair is evaluated by 5 different human evaluators, similar to \citet{koh2021text}. Since human evaluators may be unfamiliar with the distortion characteristics of 360\degree{} equirectangular images, for each pair evaluators are shown a random perspective projection from the generated images with 102\degree{} horizontal field-of-view and 16:9 aspect ratio. Images generated by our model are preferred over Pathdreamer images in 60.4\% of cases (see \cref{fig:mp3d_results}). For images with unanimous preference (preferred by 5/5 raters), our model is greatly preferred: 30.7\% compared to 12.3\% for Pathdreamer.

\subsection{Ablation Studies} \label{sec:ablations}

To validate the design choices detailed earlier, we perform an extensive ablation study summarized in \cref{table:ablation}.

\paragraph{Loss functions.} We find that the adversarial loss and $L_1$ depth reconstruction loss are both essential to strong performance: dropping the $L_1$ loss (row 2) reduces generation quality for 1 step ahead (FID@1 increases from 19.6 to 22.7), as well as over longer horizons (FID@\{1-6\} increases from 58.0 to 64.8). When we include the VGG perceptual loss~\cite{johnson2016perceptual} (row 3), we find that FID@1 is improved (19.6 to 18.2), but FID@\{1-6\} deteoriates (58.0 to 66.5). This is likely because visual quality is improved at the cost of RGB and depth consistency, which is essential for rollouts over multiple prediction steps. When KLD loss for stochastic noise conditioning~\cite{denton2018stochastic,koh2021pathdreamer} is included (row 4), generation quality worsens slightly for both 1 step and 1-6 steps ahead (FID@1 from 19.6 to 22.6, and FID@\{1-6\} from 58.0 to 65.2). While modeling stochasticity can be beneficial, we leave this out to maximize generation quality. Notably, even with the inclusion of the KLD loss and noise conditioning our results still outperform Pathdreamer, particularly in 1 step predictions (FID@1 of 22.6 vs. 27.2).

\begin{table}[t!]
\begin{minipage}{0.58\linewidth}
\begin{center}
\small
\setlength\tabcolsep{3pt}
\resizebox{0.99\linewidth}{!}{%
\begin{tabular}{@{}llcc@{}}
\noalign{\smallskip}
 \cmidrule(l){3-4} 
 \textbf{Row} & \textbf{Ablation} & 1 Step & 1--6 Steps \\ \midrule
 1 & Ours & 19.6 & 58.0 \\
 2 & $-$ $L_1$ depth loss & 22.7 & 64.8 \\
 3 & + VGG perceptual loss & 18.2 & 66.5 \\
 4 & + KLD for noise conditioning & 22.6 & 65.2 \\
 5 & + Shared decoder & 24.1 & 56.9 \\
 6 & $-$ Random masking and conv layers & 24.2 & 68.0 \\
 7 & + Ground-truth depth in GAN loss & 24.3 & 66.1 \\
\bottomrule
\end{tabular}
}
\caption{Ablations on Matterport3D (FID on Val-Unseen), using 1 context image.
}
\label{table:ablation}
\end{center}
\end{minipage}
\hspace{0.01\linewidth}%
\begin{minipage}{0.40\linewidth}
\centering
\includegraphics[width=1\linewidth]{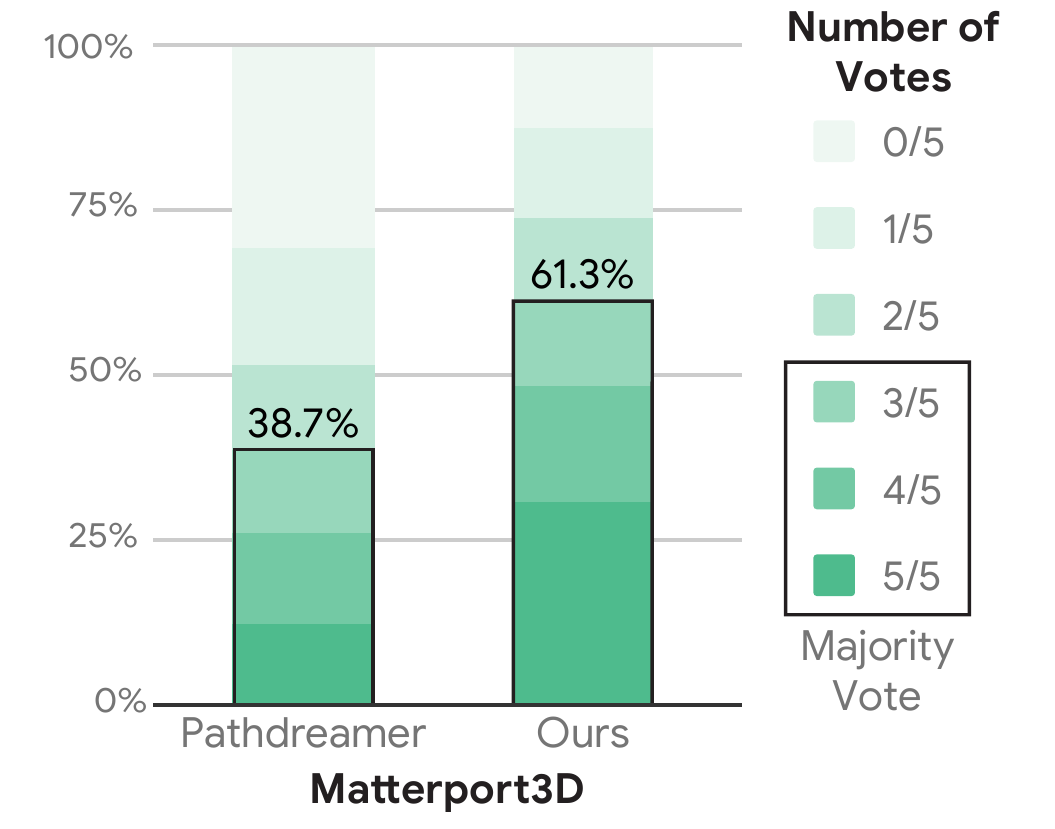}
\captionof{figure}{Human evaluations ($\uparrow$) on MP3D comparing our model to Pathdreamer.}
\label{fig:mp3d_results}
\end{minipage}
\end{table}

\paragraph{Separate decoders.} As described previously, we use a shared encoder with two separate decoders for RGB and depth. Using a single shared decoder (row  5) degrades generation quality for 1 step (FID@1 deteriorates from 19.6 to 24.1), although longer rollouts see slight improvement, improving FID@\{1-6\} from 58.0 to 56.9. This is likely due to marginally improved consistency between the RGB and depth outputs, which is essential for multistep prediction.

\paragraph{Random masking and conv layers.} We insert 4 additional $3\times3$ convolutional layers between the encoder and the decoders to increase the model's receptive field and better propagate global image information. This is combined with random masking of upto 75\% guidance pixels during training. These changes significantly improve generation quality both in one-step predictions (FID@1 improves from 24.2 in row 6 to 19.6) and over multiple steps (FID@\{1-6\} improves from 68.0 to 58.0).

\paragraph{Ground-truth depth for GAN loss.} During training, the adversarial GAN~\cite{goodfellow2014generative} hinge loss~\cite{lim2017geometric} is computed on the generated RGB-D image. We experimented with replacing the generated depth channel with the ground-truth depth channel, to explore whether this would help enforce RGB and depth consistency. Our findings suggest that it does not, with FID@1 degrading from 19.6 to 24.3 and FID@\{1-6\} from 58.0 to 66.1.

\begin{figure*}[t]
    \centering
    \includegraphics[trim={0cm 0.4cm 0 0},clip,width=1.0\linewidth]{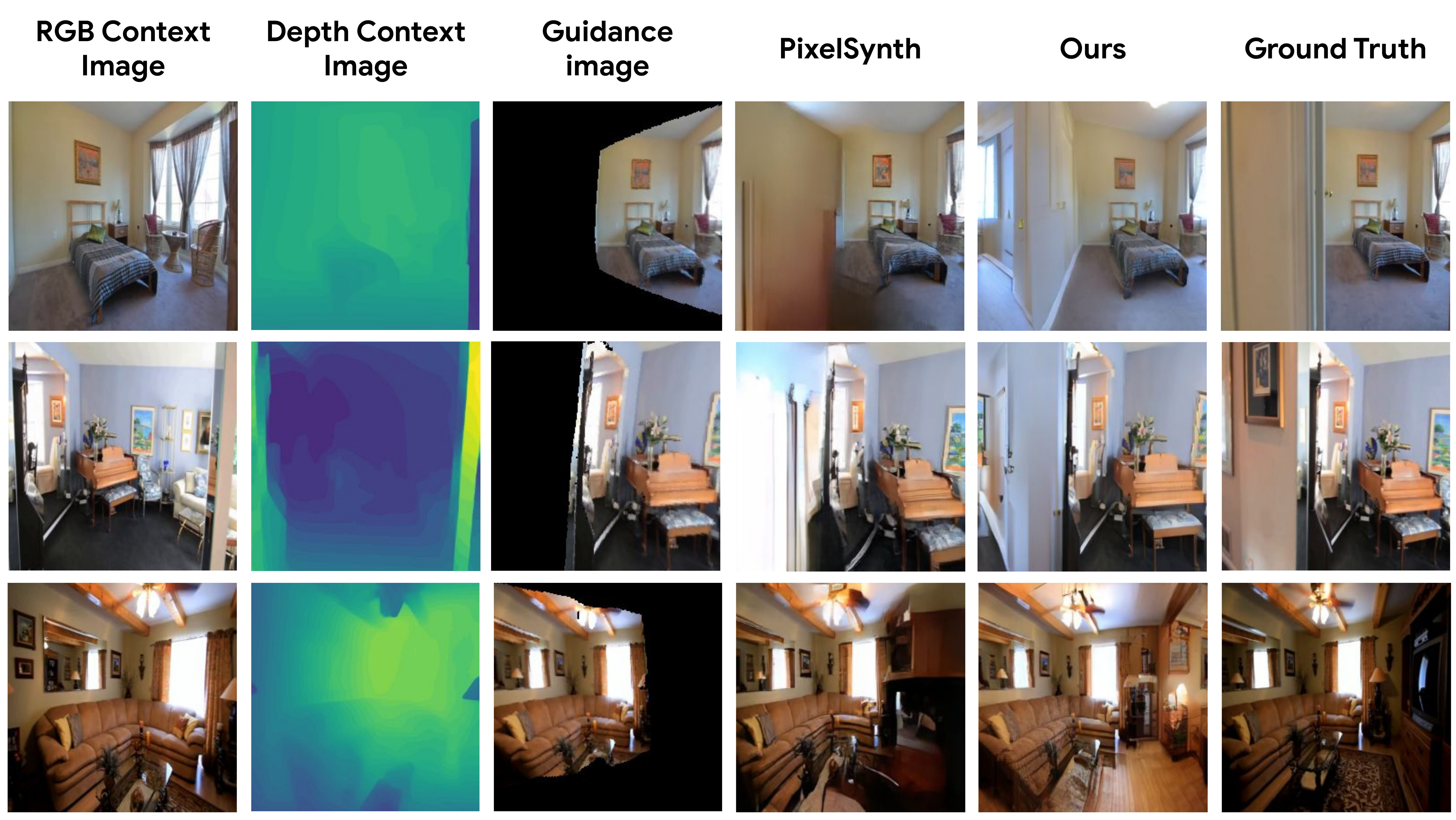}\\
    \caption{Comparison of predictions on the RealEstate10K~\cite{zhou2018stereo} dataset. In these selected examples, our model completes the scene by imagining adjacent rooms (Row 1, Row 2), while keeping wall and carpet colors consistent (Row 1)
    .}
    \label{fig:re10k_qualitative}
\end{figure*}

\subsection{RealEstate10K Experiments}
\label{sec:re10k_results}

\paragraph{Training.} The RealEstate10K~\cite{zhou2018stereo} dataset consists of a collection of real estate walkthrough videos. In its raw form, the dataset lacks depth images and camera poses, which are required to create point clouds and re-project guidance images. The PixelSynth~\cite{rockwell2021pixelsynth} and SynSin~\cite{wiles2020synsin} models, which we compare to, include a depth estimation module which is trained on the RE10K dataset using reprojection losses. We use MiDaS~\cite{Ranftl2021}, a pretrained transformer-based monocular depth estimation model, and we do not finetune on RE10K. 
See Appendix for more details.
As with the Matterport3D experiments, we train our model for single-step prediction, using a perspective camera projection for the guidance images rather than the equirectangular camera model used in the Matterport3D experiments. 
Following PixelSynth, we select image pairs for training with a camera rotation of $20\degree-60\degree$ that are estimated to be $\leq 1m$ apart, and train and evaluate with an image resolution of $256 \times 256$.

\paragraph{Evaluation.} Given an RGB context image and a target camera pose, the model must generate an RGB image for the target pose. We evaluate using the same 3,600 context-target image pairs and camera poses as PixelSynth. Since the scaling of the camera poses is arbitrary, during evaluation we scale the MiDaS depth predictions to match the PixelSynth depth predictions as closely as possible, so that the depth predictions and camera poses have consistent scaling. Qualitatively, the guidance images used by each model are extremely similar. We verify that none of the videos contributing evaluation images are in our training set.

\paragraph{FID Scores.} As reported in \cref{table:re10k_results}, our model outperforms existing methods, achieving an FID score of 23.5 compared to 25.5 and 34.7 achieved by PixelSynth ~\cite{rockwell2021pixelsynth} and SynSin~\cite{wiles2020synsin} respectively. Notably, PixelSynth results were achieved by generating 50 sample target images for each input example, and ranking them according to a combination of discriminator loss and the entropy of a scene classifier trained on MIT Places 365~\cite{Zhou2018PlacesA1} to select the best. Our results represent a single prediction from our model. Since we excluded outdoor scenes from our training data, we also report results on a subset of only indoor images (3,122 of the 3,600 images, based on manual inspection). On this subset, we achieve an FID score of 21.5 vs. 23.6 from PixelSynth. 

\begin{table}[t!]
\begin{minipage}{0.50\linewidth}
\begin{center}
\small
\setlength\tabcolsep{3pt}
\resizebox{0.90\linewidth}{!}{%
\begin{tabular}{l@{\hspace{4mm}}cc} \\
 & \multicolumn{2}{c}{\textbf{RE10K FID $\downarrow$}} \\
\cmidrule{2-3}
\textbf{Method} & All & Indoors \\ \midrule
SynSin & 34.7 & - \\
PixelSynth & 25.5 & 23.6 \\
Ours & \textbf{23.5} & \textbf{21.5} \\
\bottomrule
\end{tabular}
}
\end{center}

\caption{\footnotesize FID scores ($\downarrow$) over the RE10K val set and a subset containing only indoor images.}
\label{table:re10k_results}
\end{minipage}
\hspace{0.01\linewidth}%
\begin{minipage}{0.48\linewidth}
\centering
\includegraphics[width=1.1\linewidth]{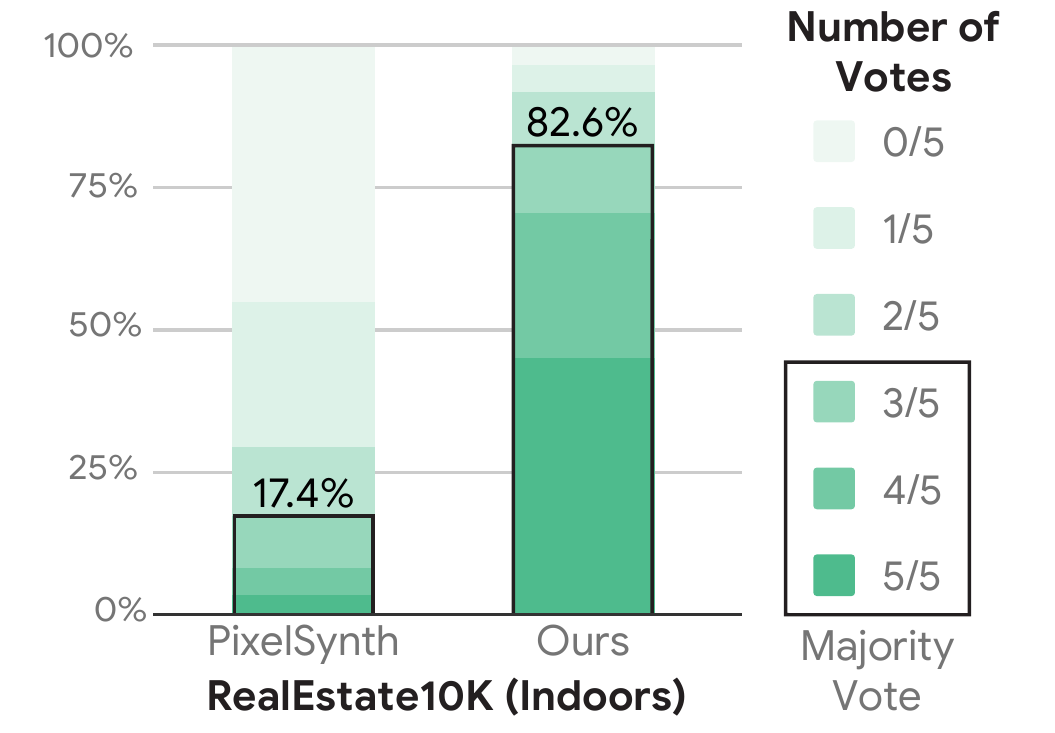}
\captionof{figure}{\footnotesize Human evaluations ($\uparrow$) comparing our model against PixelSynth on RE10K.}
\label{fig:re10k_fid}
\end{minipage}
\end{table}

\paragraph{Human Evaluations.} Similar to Matterport3D, we perform human evaluations on RE10K. As shown in \cref{fig:re10k_fid}, human annotators significantly prefer images generated by our model over PixelSynth, rating them as more realistic in 77.3\% of cases. In addition, for images which achieve unanimous preference (selected by 5/5 raters), our model is massively preferred, spanning 40.5\% of images compared to just 3.5\% for PixelSynth. Qualitative comparisons in \cref{fig:re10k_qualitative} show that our model completes some of the scenes by imagining adjacent rooms (Row 1, Row 2), while keeping wall and carpet colors consistent (Row 1), and introducing new elements like lamps and a wall painting (Row 3).

\section{Trajectory Augmentation in VLN} \label{sec:vln_results}

Motivated by our strong generation results, we investigate the usefulness of our model for synthetic \textit{trajectory augmentation}, i.e., augmenting the training data for a vision-based navigation robot by spatially perturbing training trajectories \textit{post hoc}. We focus on vision-and-language navigation (VLN) using the Room-to-Room (R2R) dataset~\cite{anderson2018vision}. This task requires an agent to follow natural language navigation instructions in unseen photorealistic indoor environments, by navigating between locations where high-res 360\degree{} Matterport3D panos have been captured. Training data consists of 14K instruction-trajectory pairs, where each trajectory is defined by a sequence of panos that are adjacent in a navigation graph. Constrained by the location of captured images, VLN agents tend to overfit to the incidental details of these trajectories~\cite{zhang2020diagnosing}, which contributes to the large performance drop in unseen environments (refer \cref{table:vln_sota_results}). We hypothesize that spatially perturbing the location of the captured images could reduce overfitting and improve generalization. 

\begin{figure*}[t!]
    \centering
    \includegraphics[width=1.0\linewidth]{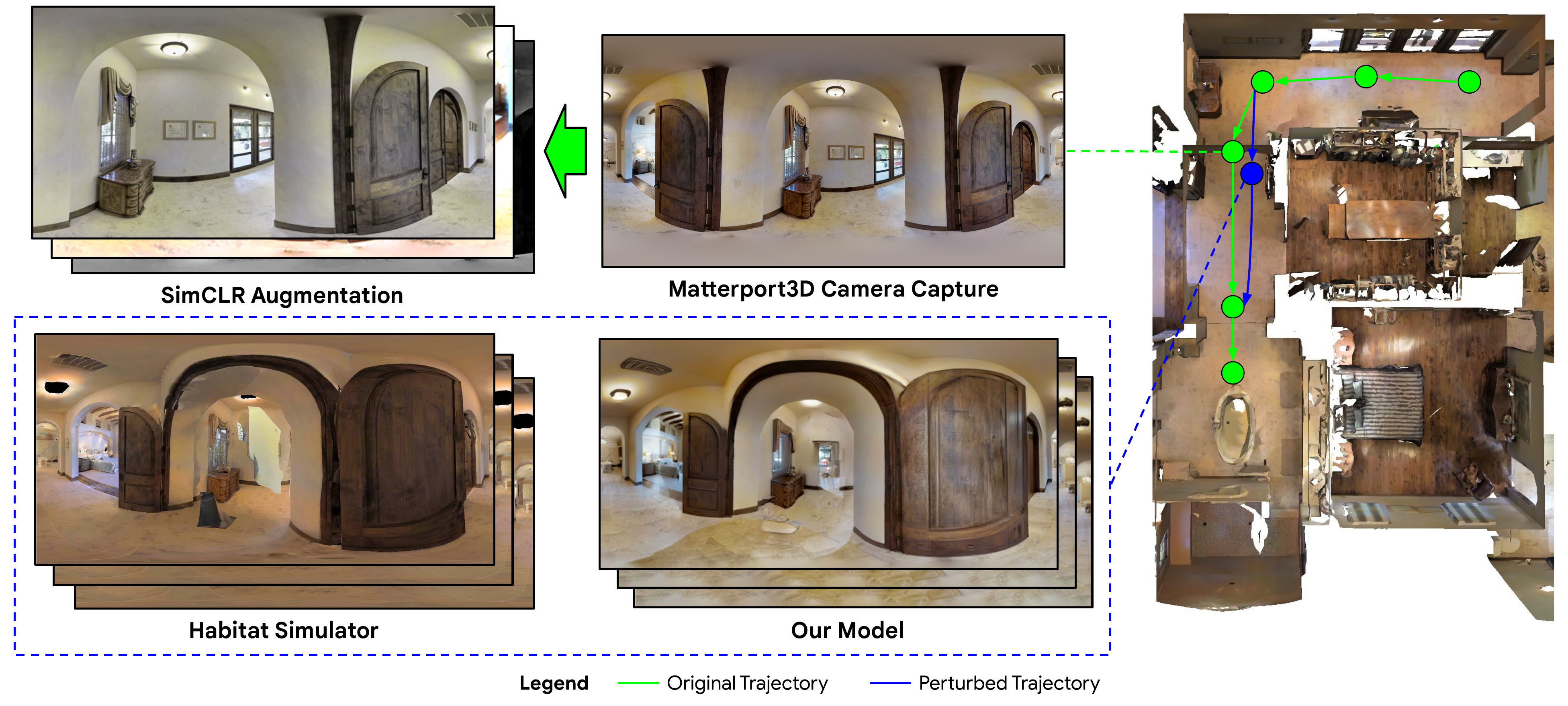} 
    \caption{Comparing SimCLR image augmentations to \textit{trajectory augmentation} using our model and/or the Habitat simulator.}
    \label{fig:augmentation}
\end{figure*}

\subsection{VLN Agent and Metrics} We base our experiments on the \vlnbert{} agent~\cite{hong2020recurrent}, an image-text cross-modal transformer with a recurrent state that is updated over time as the agent moves. The agent is trained using a mixture of imitation learning and A2C~\cite{mnih2016asynchronous}. To make the baseline as strong as possible, we first upgrade the image representation used by the model from ResNet-152~\cite{he2016deep} trained on Places365~\cite{Zhou2018PlacesA1} to MURAL-large~\cite{mural:21}, an EfficientNet-B7 architecture~\cite{efficientnet} trained on 1.8B web image-text pairs. As illustrated in \cref{table:vln_aug_results}, this improves the agent's Success Rate (SR $\uparrow$) on the R2R Val-Unseen split from 62.2\% to 67.3\%, which beats even the recent \textit{state-of-the-art} HAMT model~\cite{chen2021hamt}. SR is defined as the proportion of trajectories ending within 3m of the end of the target location. We also report Navigation Error (NE $\downarrow$), the average distance in meters between the agent's final position and the target, and Success rate weighted by the normalized inverse of the Path Length (SPL $\uparrow$)~\cite{anderson2018evaluation}.

\subsection{Trajectory Augmentation} To implement synthetic trajectory augmentation, we continually re-sample pano locations while training the VLN agent. We perturb the agent's position with a translation sampled uniformly at random from $(-1.5\text{m}, 1.5\text{m})$ for directions parallel to the ground plane and from $(-0.1\text{m}, 0.1\text{m})$ for height. To avoid perturbing the pano location through a wall or inside an object, we reject any perturbation that exceeds the depth returned in that direction. The pano is generated at $1024 \times 512$ resolution by our model trained on Matterport3D, using the nearest two ground-truth RGB-D panos as context. We compare to two alternatives: (1) data augmentation using carefully tuned SimCLR~\cite{chen2020simple} operations like random cropping, color distortion and Gaussian blur operations, and (2) rendering the spatially-perturbed pano from the textured mesh using the Habitat~\cite{habitat19iccv} simulator.

\subsection{Results}

\begin{table}[t]
\begin{center}
\scriptsize
\setlength\tabcolsep{2pt}
\resizebox{\linewidth}{!}{
\begin{tabular}{llcccccaaa}
\toprule
\multirow{2}{*}{\#} & \multicolumn{1}{l}{\multirow{2}{*}{\textbf{Model}}} && \multicolumn{3}{c}{\textbf{\valseen}} && \multicolumn{3}{a}{\textbf{\valunseen}} \\
\cmidrule{4-6} \cmidrule{8-10}
& & & NE $\downarrow$   &  SR $\uparrow$    & SPL $\uparrow$ &  & NE $\downarrow$   &  SR $\uparrow$    & SPL $\uparrow$  \\
\toprule
1. & \vlnbert (retrained) &    & 3.00  & 71.2 & 66.1 & & 3.95  & 62.2  &  56.7   \\
2. & + MURAL~\cite{mural:21}  &  & \textbf{2.78}   & \textbf{73.9}  & \textbf{68.8} &  & 3.58 &  67.3  &   60.9 \\
3. & + MURAL \& Pathdreamer~\cite{koh2021pathdreamer} &  & -   & -  & - &  & 3.41 &  67.5  &   61.0 \\
4. & + MURAL \& SimCLR~\cite{chen2020simple} &    & 3.29   & 69.4  & 63.5 &    & 3.55   & 66.5  & 59.5  \\
5. & + MURAL \& Habitat &    & 3.45   & 65.1  & 60.3 &    & 3.47   & 66.8  & 60.7  \\
6. & + MURAL \& Ours  &    & 3.09   & 71.3  & 65.6 &  & 3.41  &  68.1   &  \textbf{61.8} \\
7. & + MURAL \& Habitat \& Ours  &    & 3.29   & 68.2  & 62.7 &  & \textbf{3.29}  &  \textbf{68.8}   &  61.5   \\
\bottomrule
\end{tabular}
}
\end{center}
\caption{VLN performance on R2R~\cite{anderson2018vision}  \valunseen.
Synthetic trajectories from our model can improve an already improved model by up to 1.5\% on \valunseen (Row 6, 7), while image-based data augmentations (Row 4), simulator renders (Row 5) or Pathdreamer augmentations~\cite{koh2021pathdreamer} (Row 3) alone are ineffective.}
\label{table:vln_aug_results}
\end{table}

\begin{table}[t]
\begin{center}
\scriptsize
\setlength\tabcolsep{2pt}
\resizebox{\linewidth}{!}{
\begin{tabular}{lcccccaaacaaa}
\toprule
\multicolumn{1}{l}{\multirow{2}{*}{\textbf{Model}}} && \multicolumn{3}{c}{\textbf{\valseen}} && \multicolumn{3}{a}{\textbf{\valunseen}} && \multicolumn{3}{a}{\textbf{Test-Unseen}} \\
\cmidrule{3-5} \cmidrule{7-9} \cmidrule{11-13}
& & NE $\downarrow$   &  SR $\uparrow$    & SPL $\uparrow$ &  & NE $\downarrow$   &  SR $\uparrow$    & SPL $\uparrow$ & & NE $\downarrow$   &  SR $\uparrow$    & SPL $\uparrow$   \\
\toprule
Fast-Short~\cite{ke2017tactile} &  & - & - & - & & 4.97 & 56 & 43 & & 5.14 & 54 & 41 \\
EnvDrop~\cite{tan2019learning} &  & 3.99 & 62.1 & 59 && 5.22 & 52.2 & 48 &  & 5.23 & 51 & 47 \\
PREVALENT~\cite{hao2020towards} &  & 3.67 & 69 & 65 & & 4.71 & 58 & 53 &  & 5.30 & 54 & 51 \\
RelGraph~\cite{hong2020language} &  & 3.47 & 67 & 65 & & 4.73 & 57 & 53 &  & 4.75 & 55 & 52  \\
\vlnbert~\cite{hong2020recurrent} &  & 2.90 & 72 & 68 &  & 3.93  & 63  &  57   &  & 4.09  & 63 & 57  \\
HAMT~\cite{chen2021hamt} &  & \textbf{2.51} & \textbf{76} & \textbf{72} & & \textbf{2.29} & 66 & 61 &  & 3.93 & 65 & \textbf{60}  \\
\vlnbert (MURAL) + Habitat + Ours  &    & 3.29   & 68  & 63 &  & 3.29  &  \textbf{69}   &  \textbf{62}   &   & \textbf{3.67}   & \textbf{66}  & \textbf{60} \\
\bottomrule
\end{tabular}
}
\end{center}
\caption{VLN results on R2R~\cite{anderson2018vision} comparing against best performing prior work.}
\label{table:vln_sota_results}
\end{table}

As shown in \cref{table:vln_aug_results}, trajectory augmentation using our model improves the agent's SR and SPL on Val-Unseen by ~1\% (Row 5 vs Row 2), reducing the gap between Val-Seen and Val-Unseen from ~8\% to ~4\%. In contrast, Pathdreamer augmentations (Row 3), SimCLR image-based data augmentation (Row 4), or textured mesh based renders using Habitat (Row 5) produce no improvements over the upgraded model (Row 2). Trajectory augmentation using a combination of Habitat and our model (Row 7) produced the largest improvement in SR (+1.5\%), virtually closing the gap between Val-Seen and Val-Unseen performance, and on the unseen test set this model outperforms all published prior work (\cref{table:vln_sota_results}). The R2R dataset~\cite{anderson2018vision} is a mature benchmark at this time, and such gains do not come easily. Unlike Pathdreamer~\cite{koh2021pathdreamer}, which requires integration at inference time, our visual augmentation procedure is completely training based, and this can be applied to train virtually any off-the-shelf VLN agent.

\section{Conclusion}
Synthesizing immersive 3D environments from limited images is a challenging task brought into focus by several recent works. These approaches are highly complex, with many separately trained stages and components. We propose a simple alternative: an image-to-image GAN trained with random input masking combined with other architecture changes. Perhaps surprisingly, our approach outperforms prior work in human evaluations and on FID, and its useful for generative data augmentation as well. We achieve state-of-the-art results on the R2R dataset by spatially perturbing the training images with our model, improving generalization to unseen environments.

\clearpage

\section*{Acknowledgements}
We would like to thank Noah Snavely and many others for insightful discussions during the development of this paper. We thank Chris Rockwell for helping to set up PixelSynth models. We also thank the Google ML Data Operations team for collecting human evaluations on our generated images.

\bibliography{references}

\clearpage

\appendix
\section{Appendix}

\section{Implementation Details}

\paragraph{Training Details.} All models are implemented in TensorFlow 2.0. We set loss weights $\lambda_{\text{GAN}} = 1.0$ and $\lambda_{\text{D}} = 100.0$. Spectral normalization is used for all convolutional layers in the generator and discriminator.  We train using the Adam optimizer with parameters $\beta_1 = 0.5$ and $\beta_2 = 0.999$. The learning rates for the generator and discriminator are set to $1e^{-4}$ and $4e^{-4}$ respectively. The discriminator is trained for two training steps for each training step of the generator. For evaluation, we apply exponential moving average of generator weights with decay of $0.999$. All models are trained for 300K steps with a batch size of 128, and early stopping is applied based on FID scores on the validation dataset.

\paragraph{Evaluation Details.}
For Matterport3D, we follow Pathdreamer to compute the FID score\footnote{We used \url{https://github.com/mseitzer/pytorch-fid} for computing results.} using 10,000 random samples for each prediction sequence step. Random horizontal roll and flips are applied to obtain 10,000 samples for evaluation.

For RealEstate10K, we follow PixelSynth in computing the FID score over their evaluation set of 3,600 examples. As our training data does not contain outdoor images, we extract a subset of only indoor images (3,122 out of 3,600) based on an image classifier. We classifier frames as indoors if they are tagged with any of the ‘room’, ‘floor’, ‘kitchen’, ‘table’, ‘sofa’, ‘bed’ categories. We manually inspected images which had uncertain predictions. We ran the same evaluation on this subset of 3,122 images for both our model and PixelSynth.

\section{Architectural Details}
The detailed generator and discriminator architectures for our model can be found in \cref{tab:generator} and \cref{tab:discriminator} respectively. The same architecture is used for both Matterport3D and RealEstate10K, with the exception of the input size $H \times W$ being changed to the appropriate input dimensions ($1024\times 512$ and $256\times256$ respectively). All convolutions used in the encoder ResNet-101 are partial convolutions.

\begin{table*}[tbh]
\begin{subtable}[t]{0.49\textwidth}
\centering
\resizebox{\linewidth}{!}{%
\begin{tabular}{l}
\hline
\hline
\textbf{Input:} Context RGB-D and camera poses $(I_{1:t-1}, d_{1:t-1}, P_{1:t-1})$ \\
\hline
$(I_{1:t-1}, d_{1:t-1}, P_{1:t-1}) \rightarrow \text{Reprojection to } P_t \rightarrow x_{\text{guidance}}$ \\
\hline
$x_{\text{guidance}} \rightarrow \text{ResNet-101} \rightarrow (H // 32, W // 32, 512)$ \\
\hline
$\text{BatchNorm} \rightarrow 3\times 3 \text{ Conv} \rightarrow \text{LeakyReLU} \rightarrow (H // 32, W // 32, 512)$ \\ \hline
$\text{BatchNorm} \rightarrow 3\times 3 \text{ Conv} \rightarrow \text{LeakyReLU} \rightarrow (H // 32, W // 32, 1024)$ \\ \hline
$\text{BatchNorm} \rightarrow 3\times 3 \text{ Conv} \rightarrow \text{LeakyReLU} \rightarrow (H // 32, W // 32, 512)$ \\ \hline
$\text{BatchNorm} \rightarrow 3\times 3 \text{ Conv} \rightarrow \text{LeakyReLU} \rightarrow (H // 32, W // 32, 512)$ \\ \hline
$\text{BatchNorm} \rightarrow 3\times 3 \text{ Conv} \rightarrow h_1, (H // 32, W // 32, 512)$ \\ \hline
$h_1 \rightarrow \text{TransposedResNet-101}_{\text{D}} \rightarrow (H, W, 128)$ \\ \hline
$\text{BatchNorm} \rightarrow 3\times 3 \text{ Conv} \rightarrow \text{LeakyReLU} \rightarrow (H, W, 128)$ \\ \hline
$\text{BatchNorm} \rightarrow 3\times 3 \text{ Conv} \rightarrow \text{LeakyReLU} \rightarrow (H, W, 128)$ \\ \hline
$\text{BatchNorm} \rightarrow 3\times 3 \text{ Conv} \rightarrow \hat{d}_t, (H, W, 1)$ \\ \hline

$h_1 \rightarrow \text{TransposedResNet-101}_{\text{RGB}} \rightarrow (H, W, 128)$ \\ \hline
$\text{BatchNorm} \rightarrow 3\times 3 \text{ Conv} \rightarrow \text{LeakyReLU} \rightarrow (H, W, 128)$ \\ \hline
$\text{BatchNorm} \rightarrow 3\times 3 \text{ Conv} \rightarrow \text{LeakyReLU} \rightarrow (H, W, 128)$ \\ \hline
$\text{BatchNorm} \rightarrow 3\times 3 \text{ Conv} \rightarrow \hat{I}_t, (H, W, 3)$ \\ \hline
\textbf{Final output:} $\hat{I}_t, (H, W, 3), \hat{d}_t, (H, W, 1)$ \\ \hline
\hline
\end{tabular}
}
\caption{Generator} 
\label{tab:generator}
\end{subtable}
\hspace{\fill}
\begin{subtable}[t]{0.49\textwidth}
\centering
\resizebox{\linewidth}{!}{%
\begin{tabular}{l}
\hline
\hline
\textbf{Input:} $x =$ real/fake RGB-D images $[I_t, d_t] \in \mathbb{R}^{H\times W\times 4}$ \\
\hline

$x \rightarrow 4\times 4 \text{ Conv} \rightarrow \text{LeakyReLU} \rightarrow (H/2, W/2, 128) $\\
\hline
$4\times 4 \text{ Conv} \rightarrow \text{InstanceNorm} \rightarrow \text{LeakyReLU} \rightarrow (H/4, W/4, 256) $\\  
\hline
$4\times 4 \text{ Conv} \rightarrow \text{InstanceNorm} \rightarrow \text{LeakyReLU} \rightarrow (H/8, W/8, 512) $\\  
\hline
$4\times 4 \text{ Conv} \rightarrow \text{InstanceNorm} \rightarrow \text{LeakyReLU} \rightarrow (H/16, W/16, 512) $\\  
\hline
$4\times 4 \text{ Conv} \rightarrow \text{InstanceNorm} \rightarrow \text{LeakyReLU} \rightarrow (H/32, W/32, 512) $\\  
\hline
$4\times 4 \text{ Conv} \rightarrow \text{InstanceNorm} \rightarrow \text{LeakyReLU} \rightarrow (H/32, W/32, 512) $\\  
\hline
$4\times 4 \text{ Conv} \rightarrow Out1, (H/32, W/32, 1)$ \\ 
\hline

$\text{x} \rightarrow 3 \times 3 \text{ stride } 2 \text{ AvgPool} \rightarrow x_1, (H/2, W/2, 4)$ \\
\hline

$x_1 \rightarrow 4\times 4 \text{ Conv} \rightarrow \text{LeakyReLU} \rightarrow (H/4, W/4, 128) $\\
\hline
$4\times 4 \text{ Conv} \rightarrow \text{InstanceNorm} \rightarrow \text{LeakyReLU} \rightarrow (H/8, W/8, 256) $\\  
\hline
$4\times 4 \text{ Conv} \rightarrow \text{InstanceNorm} \rightarrow \text{LeakyReLU} \rightarrow (H/16, W/16, 512) $\\  
\hline
$4\times 4 \text{ Conv} \rightarrow \text{InstanceNorm} \rightarrow \text{LeakyReLU} \rightarrow (H/32, W/32, 512) $\\  
\hline
$4\times 4 \text{ Conv} \rightarrow \text{InstanceNorm} \rightarrow \text{LeakyReLU} \rightarrow (H/64, W/64, 512) $\\  
\hline
$4\times 4 \text{ Conv} \rightarrow \text{InstanceNorm} \rightarrow \text{LeakyReLU} \rightarrow (H/64, W/64, 512) $\\  
\hline
$4\times 4 \text{ Conv} \rightarrow Out2, (H/64, W/64, 1)$ \\ 
\hline
\textbf{Final output:} $(Out1 + Out2) / 2$ \\
\hline
\hline
\end{tabular}
}
\caption{Discriminator}
\label{tab:discriminator}
\end{subtable}
\caption{Generator and discriminator architectures.} 
\label{tab:architecture}
\end{table*}

\section{RealEstate10K Experiment details}

\section{COLMAP sampling details}
Now, we describe the details of our RealEstate10K training dataset. We obtain camera trajectories and depth maps for Youtube videos listed by running SLAM and bundle adjustment on the video frames. RealEstate10K breaks down the YouTube video into a collection of smaller sub-sequences. Each sub-sequence consists of connected frames that were tracked by ORB-SLAM~\cite{MurArtal2015ORBSLAMAV}. Out of the original dataset, only $6,865$ videos are available because the rest have been taken down since the release. We ran the original pipeline from RealEstate10K but with longer sequence length (up to 1,000 frames) which results in a total of $108,275$ sub-sequences containing over $19$ million frames. Videos in the RealEstate10K dataset often contain walkthroughs of outdoor surroundings as well. Since we are only interested in synthesizing indoor scenes, we filter sub-sequences which contain outdoor frames from our dataset. We filter these by running a multi-label classifier which tags each frame of the sub-sequence with relevant semantic categories. We remove those sub-sequences in which no frame is tagged with any of the `room', `floor', `kitchen', `table', `sofa', `bed' categories. After filtering, we are left with 83,559 sub-sequences containing over $15$ million frames. We obtain camera trajectories corresponding to each sub-sequence by running COLMAP~\cite{schoenberger2016sfm}, a structure-from-motion pipeline. COLMAP was able to successfully build dense reconstruction for 67,834 out of the 83,559 sub-sequences. 

To obtain per-frame depth maps, we use a standard multi-view stereo system in COLMAP. The depth estimates from MVS are often too noisy due to camera-blur, reflections, poor lightning, and other factors~\cite{Li2021MannequinChallengeLT}. Following procedure outlined in Li et al.~\cite{Li2021MannequinChallengeLT}, we first filter outlier depth values using the depth refinement method outlined in ~\cite{li2018megadepth}. Further, we remove errorneous depth values by considering the consistency between MVS depth and depth obtained from motion parallax between adjacent pair of frames. Depth estimates from MVS has another limitation. These depth estimates are often extremely sparse. This severely limit the density of point-cloud constructed from context images, and leads to sparse guidance images for the model. To overcome this, we generate dense depth predictions from MiDaS~\cite{Ranftl2021}. MiDaS is a state-of-the-art transformer based monocular depth estimation model which outputs a dense depth map from a single RGB image. Of course, the scale of the depth predictions from MiDaS for different frames are inconsistent and not aligned with the camera pose. Thus they cannot be used to create a unified point-cloud. We tackle this by a simple fix. We scale the depth predictions from MiDaS to match the COLMAP MVS depth outputs for valid pixels using the least-squares method. We select only those frames for training for which the depth map produced by MVS covers at least 10\% of the image.

\section{Human Evaluations}

\subsection{User Interface and Process}  \label{sec:human_eval_ui}
\begin{figure*}[t]
    \centering
    \includegraphics[width=0.9\textwidth]{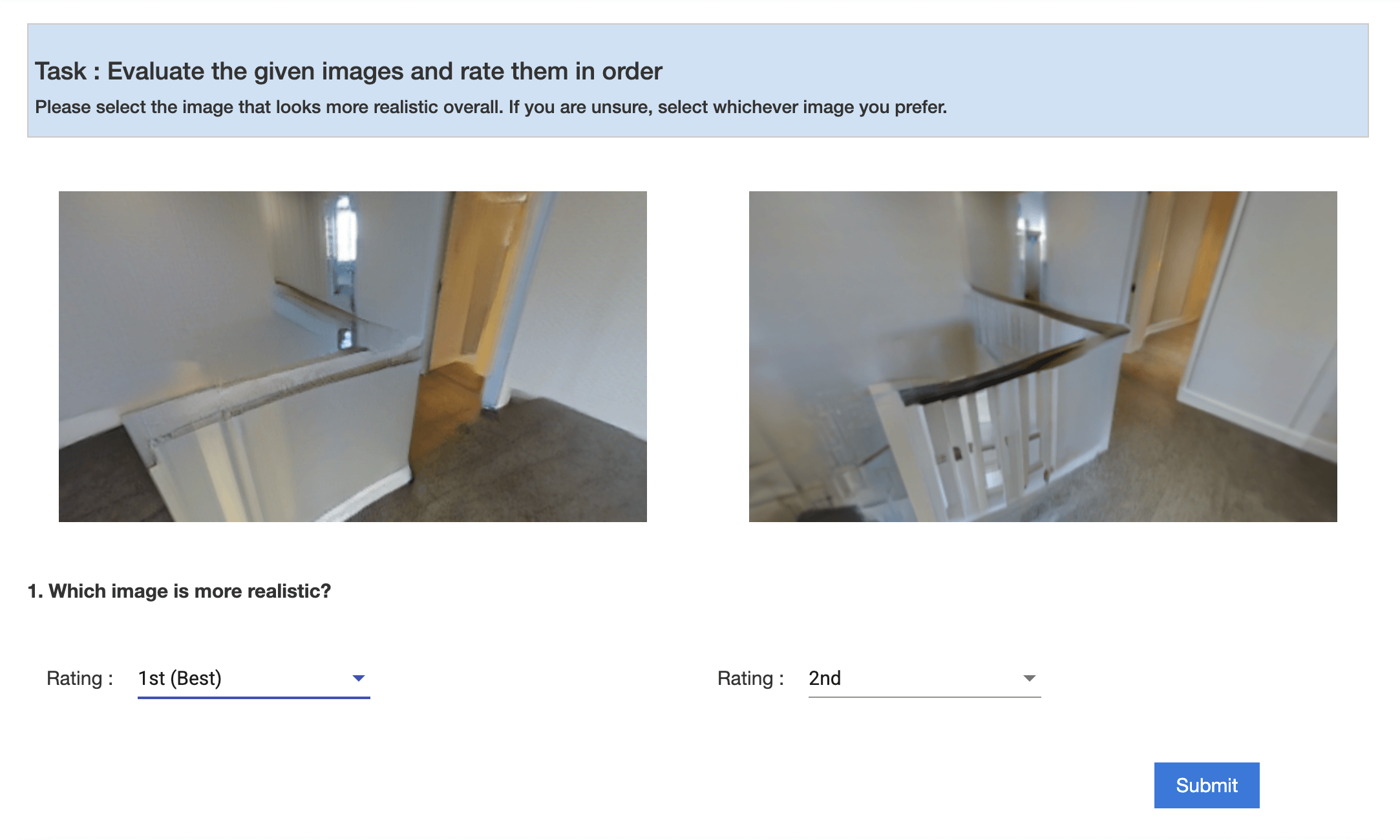}
    \caption{User interface shown to human raters for evaluating models on Matterport3D and RealEstate10K. Models are anonymized and shown in random order.}
    \label{fig:human_eval_ui}
\end{figure*}

We perform human evaluations of 1,000 image pairs for both the Matterport3D and RealEstate10K datasets. For Matterport3D, we do head-to-head comparisons between our model and Pathdreamer, and for RealEstate10K, we compare our model against PixelSynth. Each pair is evaluated by 5 independent human evaluators, and the user interface is shown in \cref{fig:human_eval_ui}. The two models are anonymized and shown in random order to minimize biases.

\subsection{Evaluation with Guidance Image}

\begin{figure*}[t]
    \centering
    \includegraphics[width=0.62\textwidth]{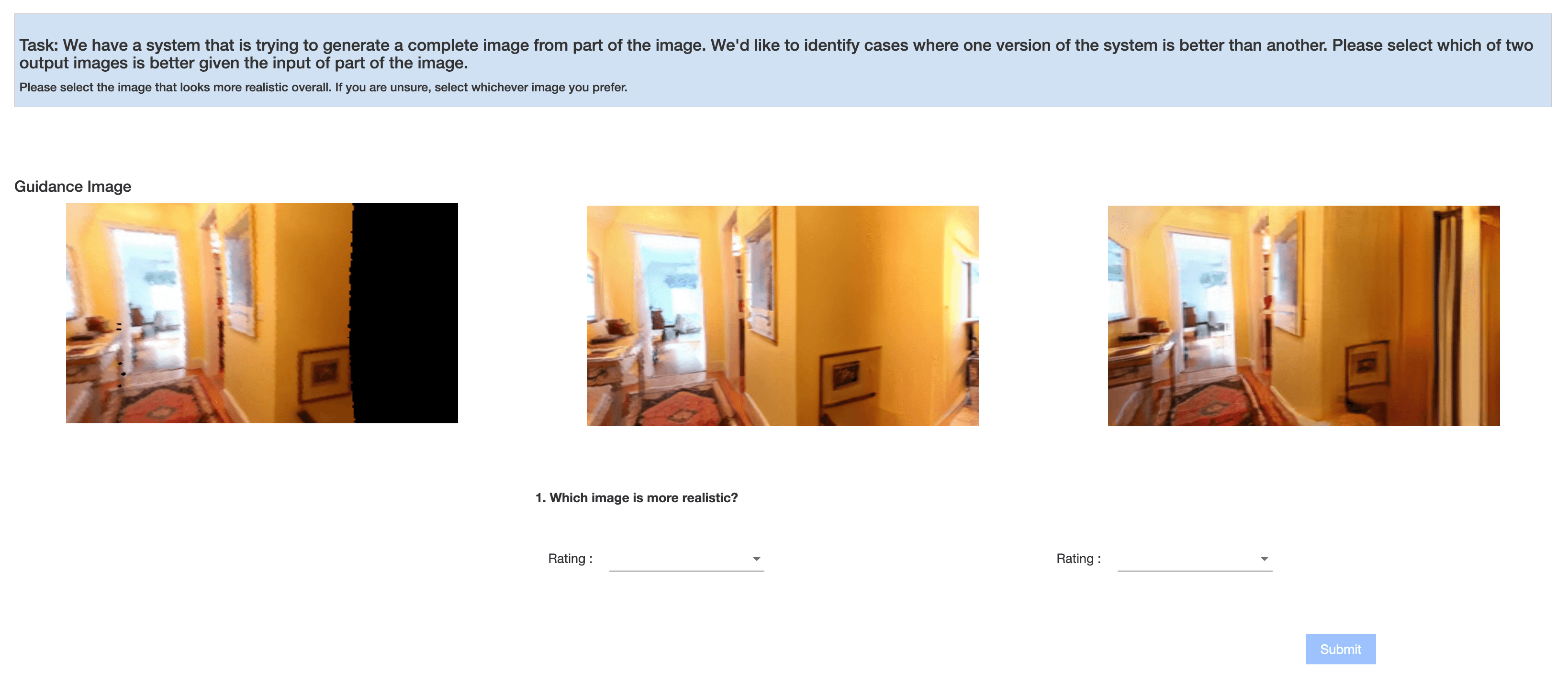}
    \includegraphics[width=0.37\textwidth]{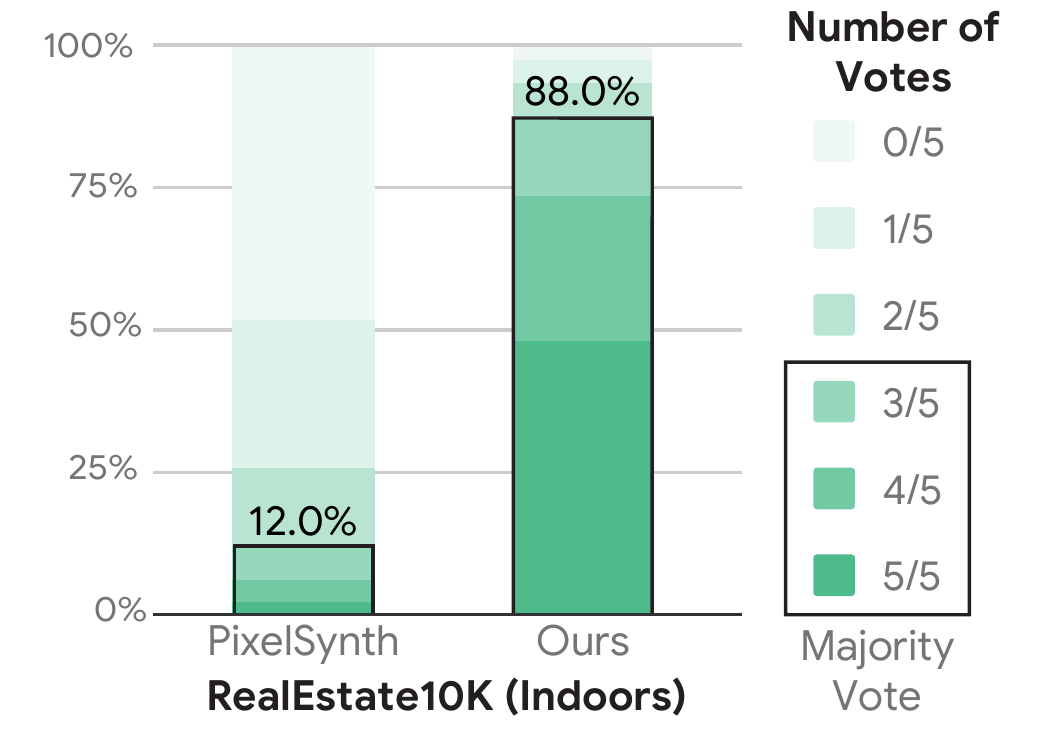}
    \caption{User interface shown to human raters for performing evaluations RealEstate10K with guidance input (left), and human evaluations ($\uparrow$) comparing our model to PixelSynth (right).}
    \label{fig:guidance_eval_ui}
\end{figure*}

In addition to the evaluation described in our main paper and in the \textit{Human Evaluations} section, we peform a secondary evaluation for RealEstate10K similar to the process in PixelSynth. In this setup, we show the input guidance image to the evaluators, and task them to select the image that best matches the partial input (see \cref{fig:guidance_eval_ui}). Under this procedure, we expect the evaluators to focus even more on the regions of the image requiring inpainting. Under this setting, our model significantly outperforms PixelSynth, with 88.0\% of examples being preferred by human evaluators (compared to 12\% for PixelSynth). For examples with unanimous preference (5/5 raters prefer it), our model is overwhelming preferred: 48.4\% compared to 2.0\% for PixelSynth.

\section{Qualitative Evaluation}

\begin{figure*}[t]
\centering
    \begin{subfigure}{1.0\textwidth}
    \includegraphics[width=1.0\linewidth]{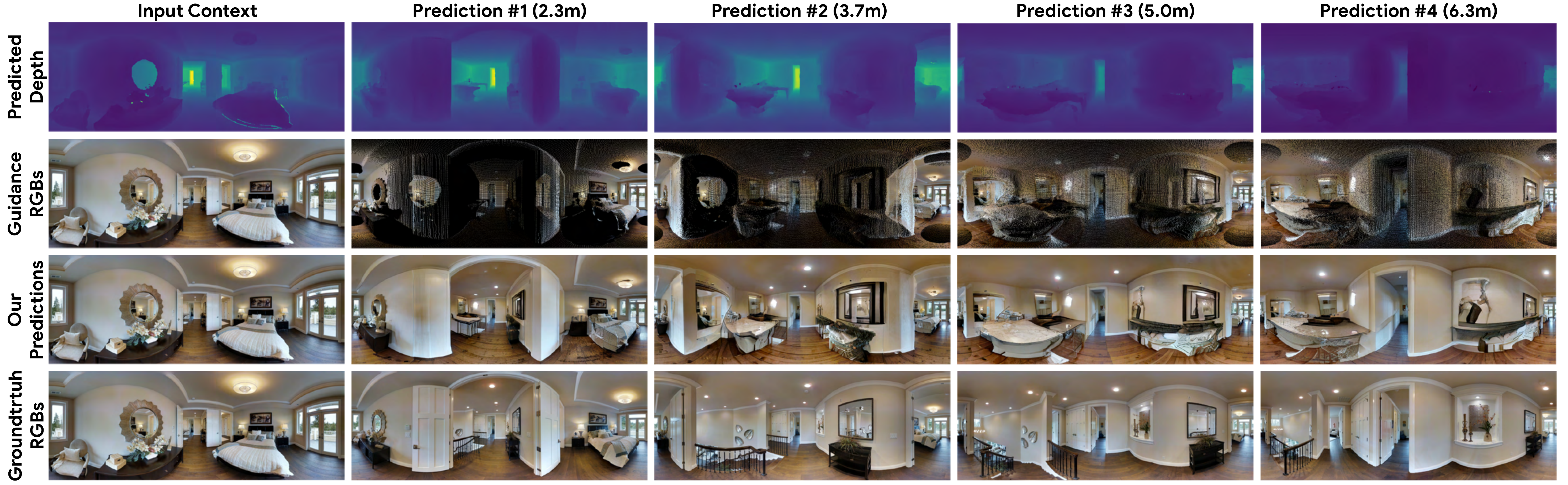}
    \caption{At 2.3m, our model infills the missing region to hallucinate a table, which remains consistent at the following step (3.7m) and beyond. Even at 6.3m away, the 3D geometry (e.g., the doorway) of the scene is well preserved.}
    \end{subfigure}
    \vspace{0.1in}

    \begin{subfigure}{1.0\textwidth}
    \includegraphics[width=1.0\linewidth]{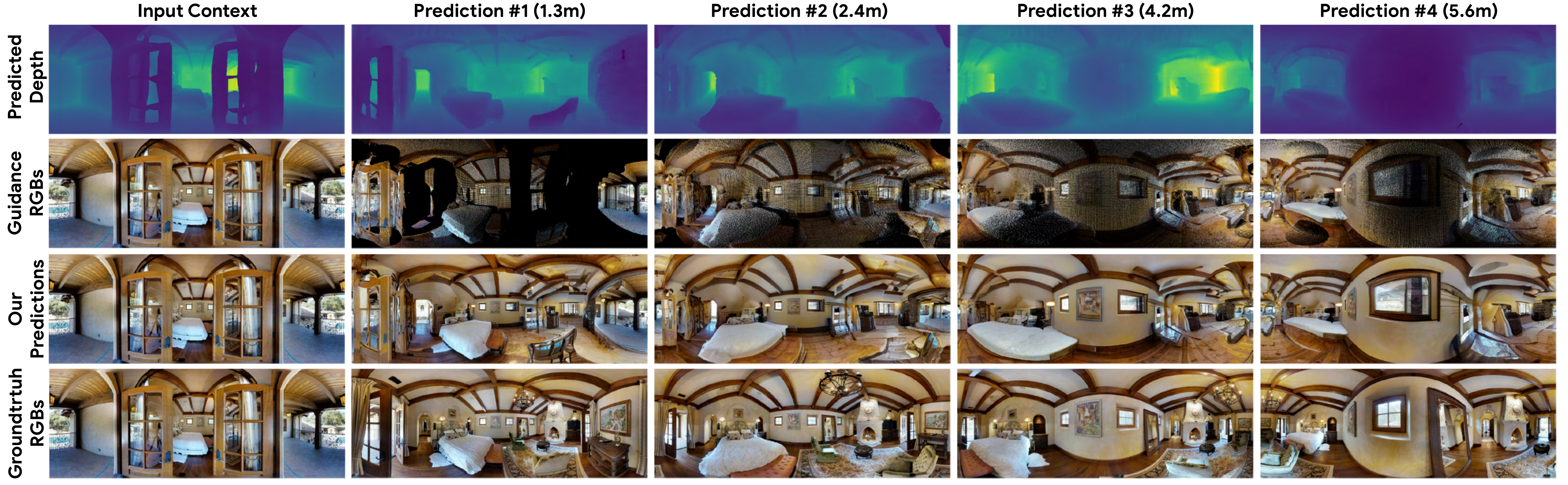}
    \caption{At 1.3m, our model provides a plausible infilling of the missing regions of the scene, enabling accurate and geometrically consistent generation at future steps. Even at 5.6m away, the room generation results remain plausible.}
    \end{subfigure}
    \vspace{0.1in}
    
    \begin{subfigure}{1.0\textwidth}
    \includegraphics[width=1.0\linewidth]{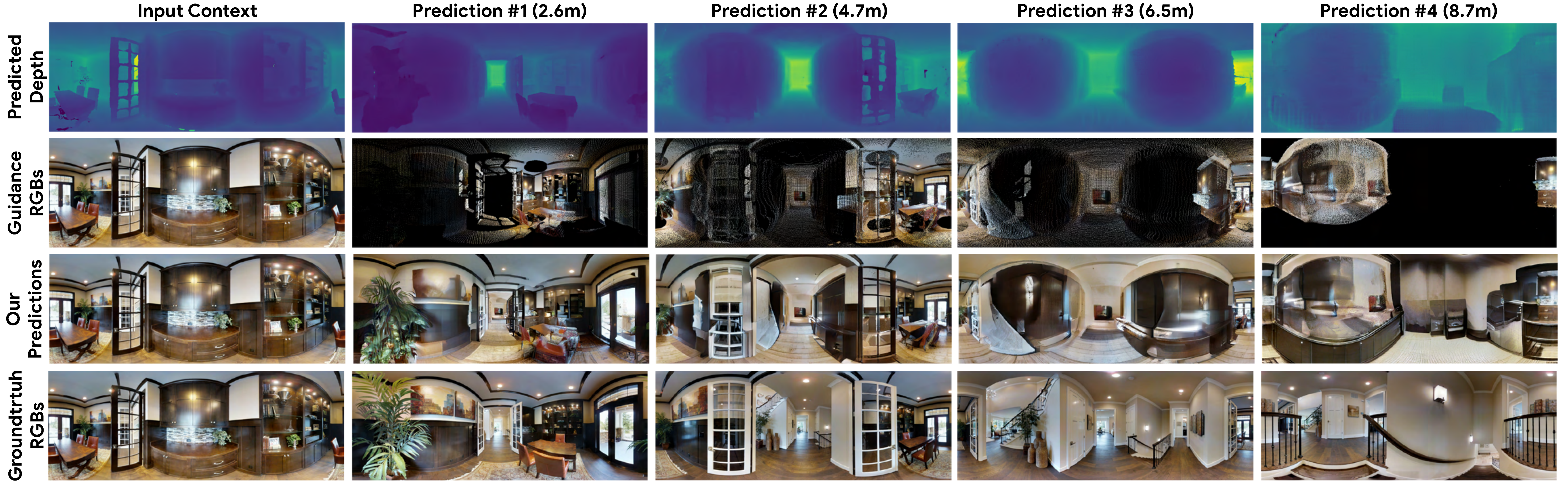}
    \caption{Our model preserves high frequency textures well, as seen by the upscaling and generation of the plant at 2.6m. Our model also correctly infills the content of the doorway in the center at 2.6m, enabling plausible predictions at 4.7m and beyond.}
    \end{subfigure}
    \vspace{0.1in}


\caption{Selected examples of predicted sequences from the Val-Unseen R2R split of Matterport3D using one ground truth observation as context.}
\label{fig:mp3d_qualitative_cherry}
\end{figure*}

\begin{figure*}[t]
\centering
    \begin{subfigure}{1.0\textwidth}
    \includegraphics[width=1.0\linewidth]{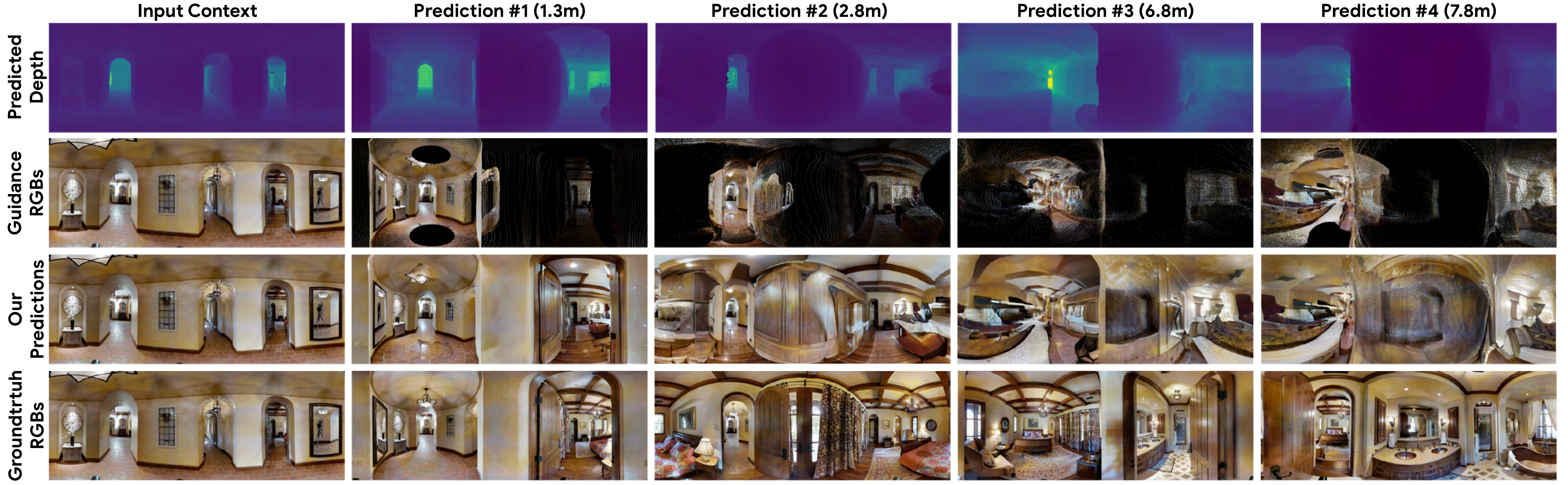}
    \caption{At 1.3m, our model infills and completes the scene, generating an image that matches the groundtruth geometry very closely. This enables simulated traversal through the scene, producing fairly plausible generation results even at 6.8m and beyond.}
    \end{subfigure}
    \vspace{0.1in}

    \begin{subfigure}{1.0\textwidth}
    \includegraphics[width=1.0\linewidth]{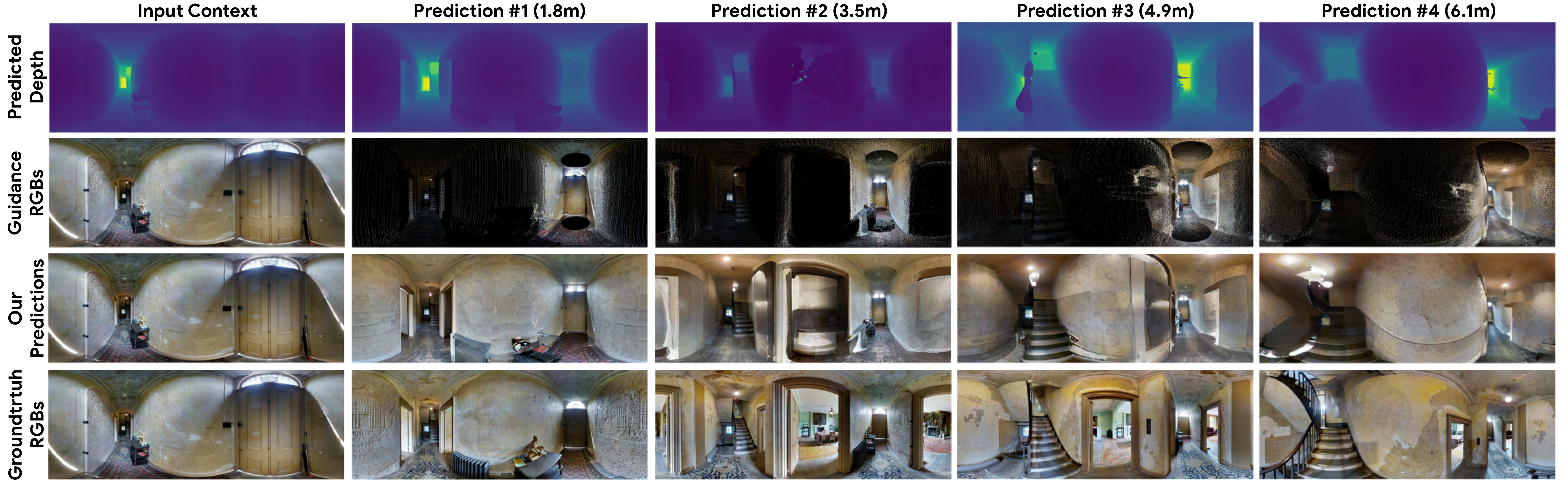}
    \caption{Our model is able to predict plausible generation results that respect the geometry of this hallway, including upsampling of the stairwell at step 3.5m. Although the model does not generate the room from the groundtruth at 4.9m, it generates a plausible result assuming a wall.}
    \end{subfigure}
    \vspace{0.1in}

    \begin{subfigure}{1.0\textwidth}
    \includegraphics[width=1.0\linewidth]{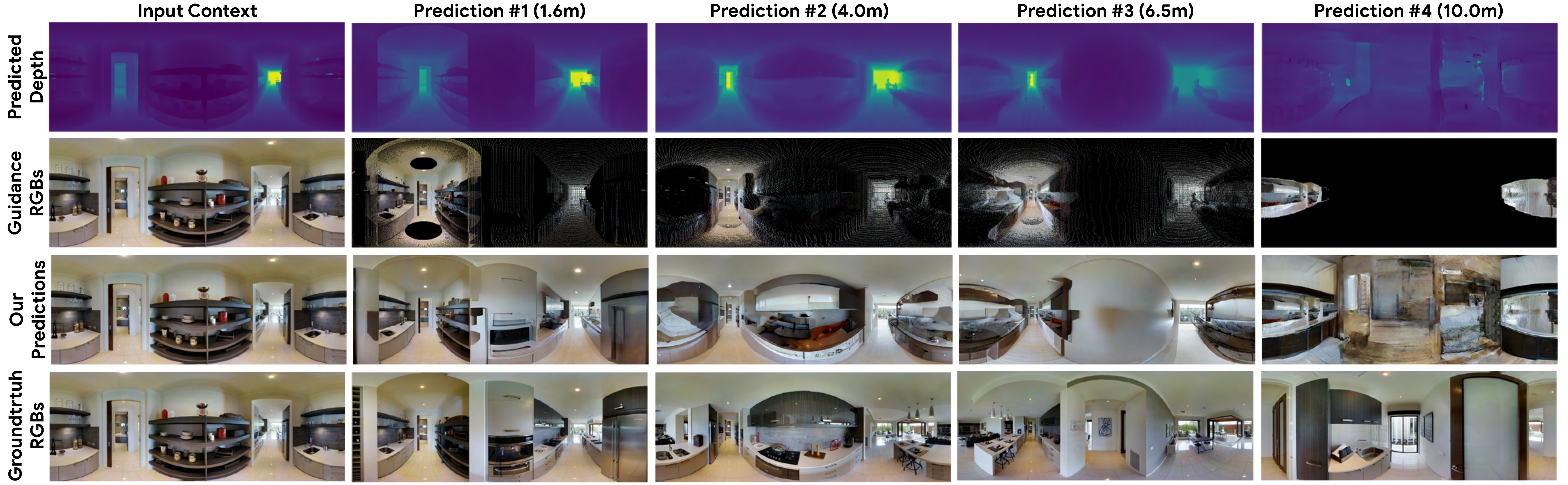}
    \caption{At 4.0m, our model infills and completes the scene, imagining and generating a complete countertop for the kitchen.}
    \end{subfigure}
    \vspace{0.1in}

\caption{Randomly sampled examples of predicted sequences from the Val-Unseen R2R split of Matterport3D using one ground truth observation as context.}
\label{fig:mp3d_qualitative_random}
\end{figure*}

\begin{figure*}[t]
\centering
    \begin{subfigure}{1.0\textwidth}
    \includegraphics[width=1.0\linewidth]{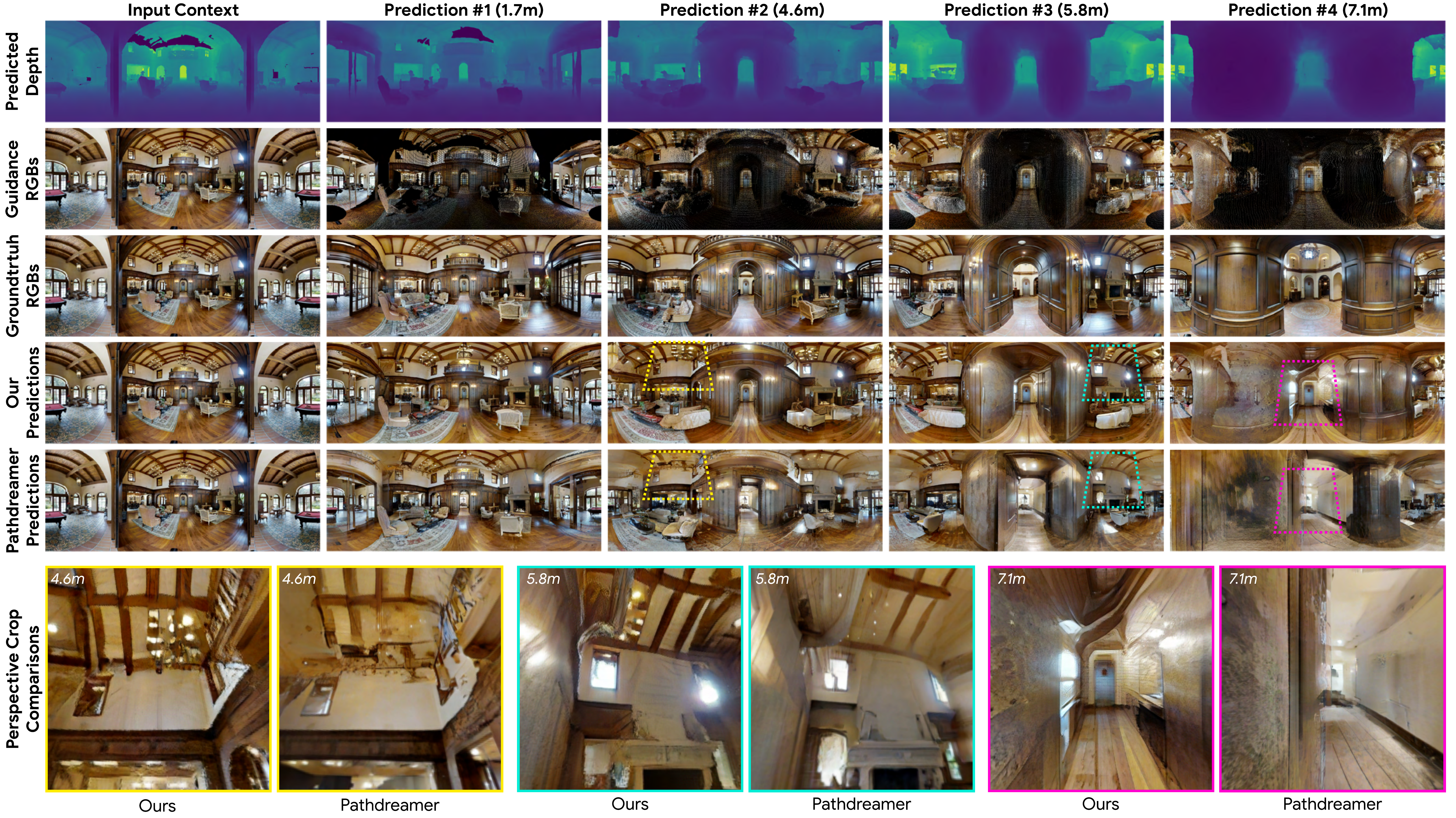}
    \caption{Our model generates higher quality ceiling beams compared to Pathdreamer at 4.6m and 5.8m. Our model is also able to replicate a portion of the ceiling lights at 4.6m, which Pathdreamer is unable to do. Our generation results produce higher quality textures in general, such as the hardwood floors and carpet at 4.6m.}
    \end{subfigure}
    \vspace{0.1in}

    \begin{subfigure}{1.0\textwidth}
    \includegraphics[width=1.0\linewidth]{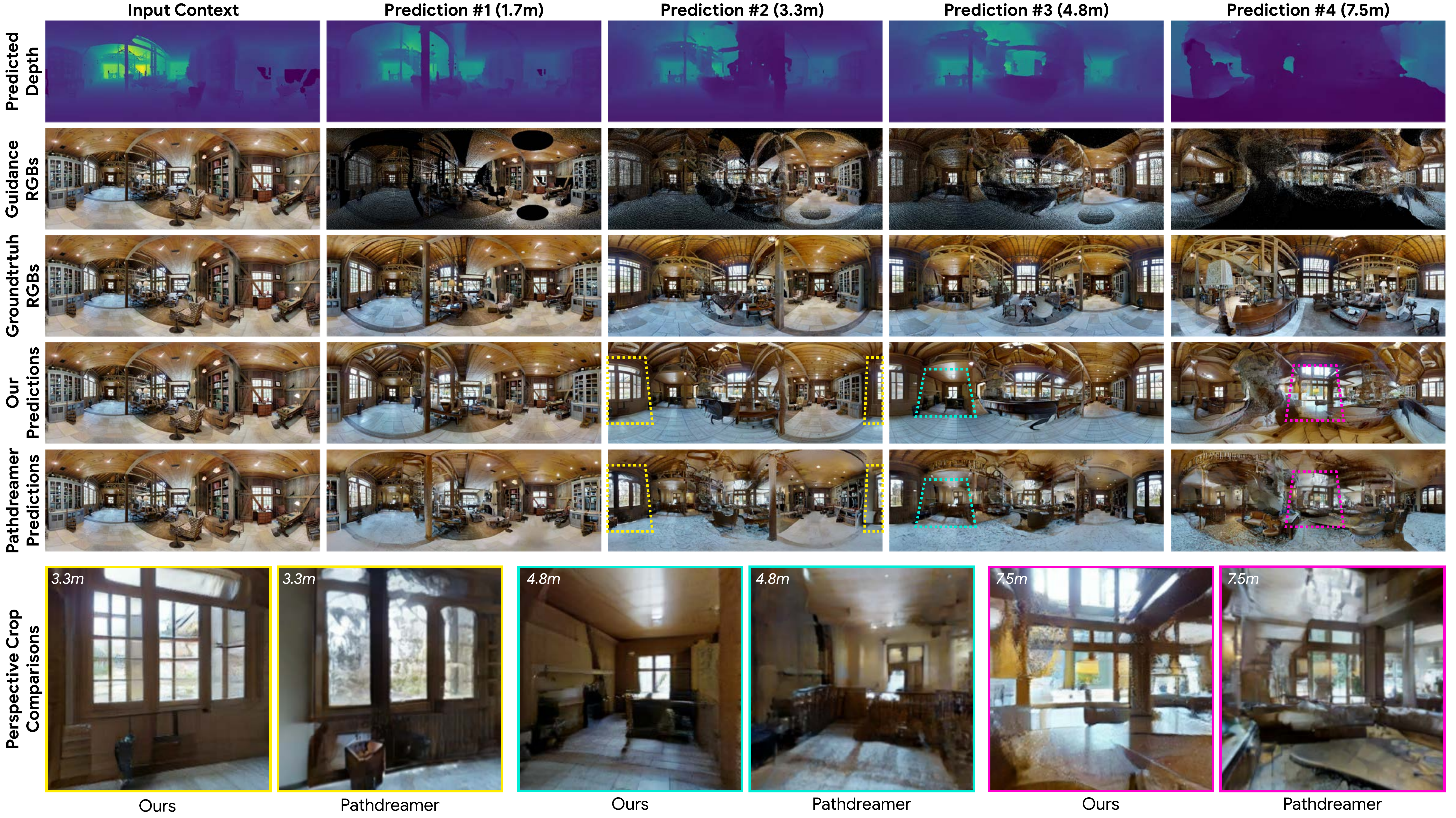}
    \caption{Our model produces clearer windows, and is able to replicate the checkerbox pattern of the window frame at 3.3m. At 4.8m, our model is able to generate a more coherent scene of a table and a window, compared to Pathdreamer which generates blurry and indistinctive outputs. Our model is also capable of generating objects more distinctly compared to Pathdreamer, such as the lamp at 7.5m.}
    \end{subfigure}
    \vspace{0.1in}

\caption{Selected examples comparing our model to Pathdreamer. One RGB-D pano is provided as context, and a sequence of novel viewpoints are generated.}
\label{fig:mp3d_qualitative_pathdreamer}
\end{figure*}

\begin{figure*}[t]
\centering
\includegraphics[width=0.9\linewidth]{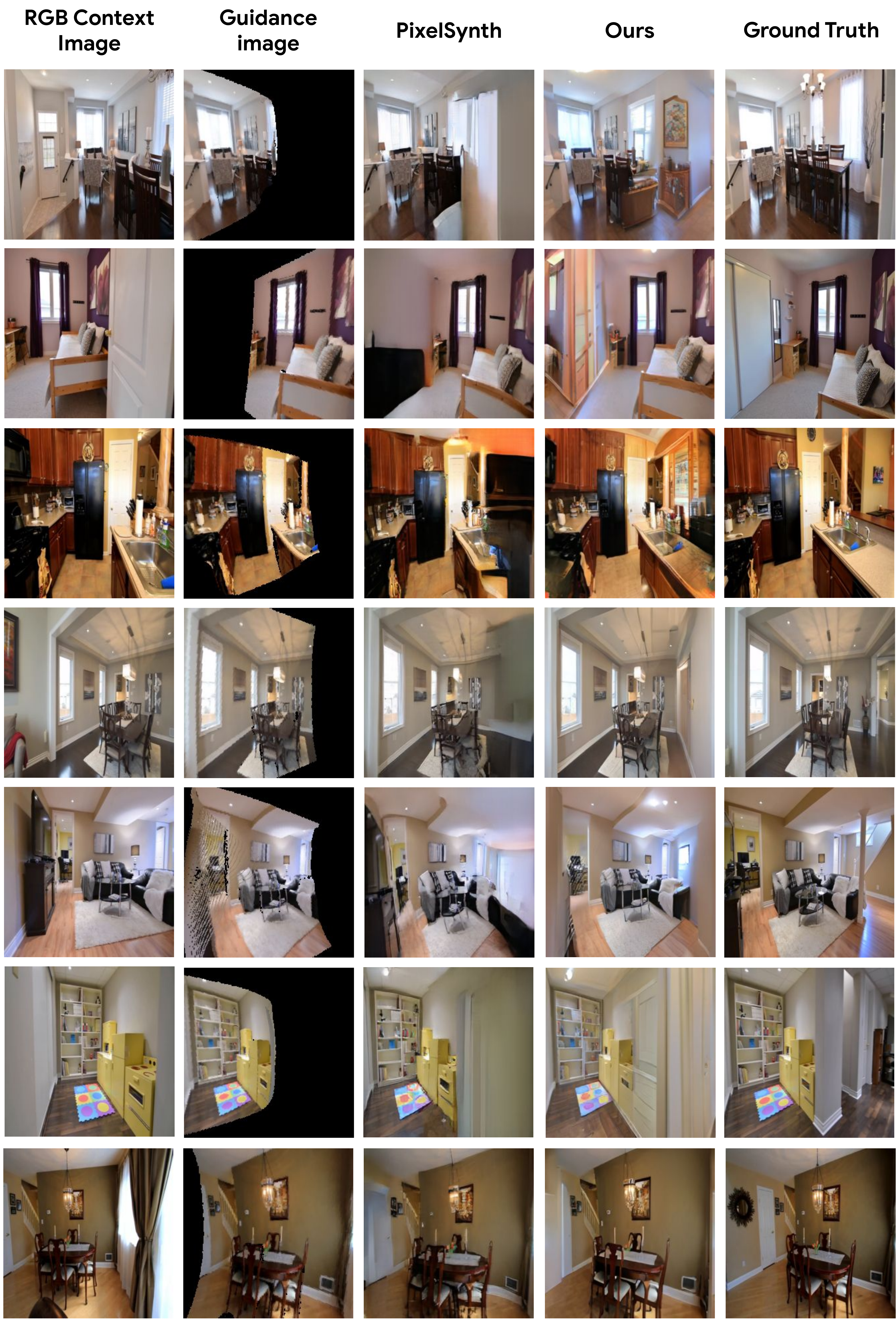}
\vspace{-0.1in}
\caption{Cherry picked examples of predicted sequences from the evaluation split of RealEstate10K using one ground truth observation as context.}
\label{fig:re10k_qualitative_cherry}
\end{figure*}

\begin{figure*}[t]
\centering
\includegraphics[width=0.9\linewidth]{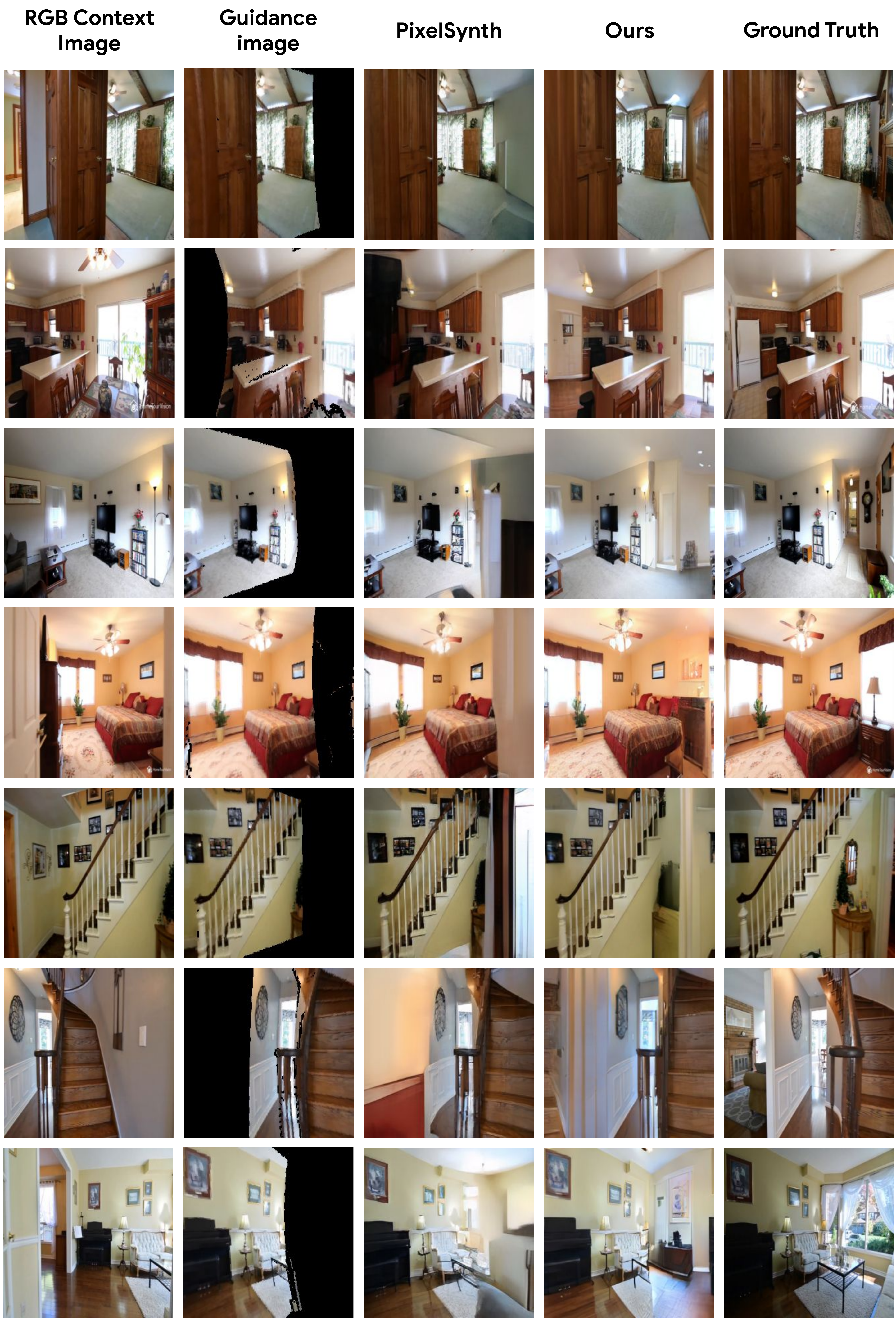}
\vspace{-0.1in}
\caption{Random sampled examples of predicted sequences from the evaluation split of RealEstate10K using one ground truth observation as context.}
\label{fig:re10k_qualitative_random}
\end{figure*}

In this section, we present additional qualitative results generated by our model. We show cherry-picked examples (\cref{fig:mp3d_qualitative_cherry}) and random examples  (\cref{fig:mp3d_qualitative_random}) on the Matterport3D dataset. Similarly, for the RealEstate10K dataset we show more cherry-picked examples (\cref{fig:re10k_qualitative_cherry}) and random examples (\cref{fig:re10k_qualitative_random}).
We also provide more side-by-side comparisons of our model against Pathdreamer in \cref{fig:mp3d_qualitative_pathdreamer}.

\subsection{Video Generation Results}

Our model is also capable of generating compelling, high resolution video by subsampling trajectories with small viewpoint changes. We provide videos displaying generation results for unseen environments in Matterport3D, and for rotations and camera trajectories from RealEstate10K. We refer readers to the video\footnote{Video results: \videourl} for these results.

\end{document}